\documentclass[lettersize,journal]{IEEEtran}
\usepackage{amsmath,amsfonts}
\usepackage{algorithm}
\usepackage{algpseudocode}
\usepackage{array}
\usepackage[caption=false,font=normalsize,labelfont=sf,textfont=sf]{subfig}
\usepackage{textcomp}
\usepackage{stfloats}
\usepackage{url}
\usepackage{verbatim}
\usepackage{graphicx}
\usepackage{cite}

\usepackage[colorlinks,
            linkcolor=blue,
            anchorcolor=blue,
            citecolor=blue,
            urlcolor=blue
            ]{hyperref}
\usepackage{xspace}

\usepackage{amssymb}
\usepackage{booktabs}
\usepackage{multirow}
\usepackage{color}
\usepackage{xcolor}
\usepackage{booktabs,colortbl}

\usepackage{pifont}
\usepackage{arydshln}
\definecolor{lightgrayv}{HTML}{F4F3F8} 
\definecolor{grayv}{HTML}{707070}
\definecolor{redv}{HTML}{C00000}

\newcommand{\eg}{\emph{e.g.,}\xspace}

\newcommand{\etc}{\emph{etc.}\xspace}

\newcommand{\baby}{\textsc{Havc-m}$^4$\textsc{d}\xspace}

\begin{document}

\title{Harmful Visual Content Manipulation Matters in Misinformation Detection Under Multimedia Scenarios}


\author{Bing Wang, Ximing Li, Changchun Li, Jinjin Chi, Tianze Li, Renchu Guan, Shengsheng Wang

\thanks{This work was supported in part by the National Science and Technology Major Project of China (No. 2021ZD0112500), the National Natural Science Foundation of China (No.62276113, No.62376106), and China Postdoctoral Science Foundation (No.2022M721321).}
\thanks{Bing Wang, Ximing Li, Changchun Li, Jinjin Chi, Renchu Guan, Shengsheng Wang are with College of Computer Science and Technology, Jilin University, China (e-mail: wangbing1416@gmail.com, liximing86@gmail.com, changchunli93@gmail.com, chijinjin616@gmail.com, guanrenchu@jlu.edu.cn, wss@jlu.edu.cn). Ximing Li is the corresponding author (\url{liximing86@gmail.com}).}
\thanks{Tianze Li is with the School of Advanced Technology, Xi'an Jiaotong-Liverpool University, China. Email: Tianze.Li22@student.xjtlu.edu.cn}}
%

\markboth{Journal of \LaTeX\ Class Files,~Vol.~14, No.~8, August~2021}%
{Shell \MakeLowercase{\textit{et al.}}: A Sample Article Using IEEEtran.cls for IEEE Journals}


\maketitle

\begin{abstract}

Nowadays, the widespread dissemination of misinformation across numerous social media platforms has led to severe negative effects on society. To address this challenge, the automatic detection of misinformation, particularly under multimedia scenarios, has gained significant attention from both academic and industrial communities, leading to the emergence of a research task known as Multimodal Misinformation Detection (MMD). Typically, current MMD approaches focus on capturing the semantic relationships and inconsistency between various modalities but often overlook certain critical indicators within multimodal content. Recent research has shown that manipulated features within visual content in social media articles serve as valuable clues for MMD. Meanwhile, we argue that the potential intentions behind the manipulation, \eg harmful and harmless, also matter in MMD. Therefore, in this study, we aim to identify such multimodal misinformation by capturing two types of features: manipulation features, which represent if visual content has been manipulated, and intention features, which assess the nature of these manipulations, distinguishing between harmful and harmless intentions. Unfortunately, the manipulation and intention labels that supervise these features to be discriminative are unknown. To address this, we introduce two weakly supervised indicators as substitutes by incorporating supplementary datasets focused on image manipulation detection and framing two different classification tasks as positive and unlabeled learning issues. With this framework, we introduce an innovative MMD approach, titled Harmful Visual Content Manipulation Matters in MMD (\baby). Comprehensive experiments conducted on four prevalent MMD datasets indicate that \baby significantly and consistently enhances the performance of existing MMD methods.

\end{abstract}


\section{Introduction}

\begin{figure}[t]
  \centering
  \includegraphics[scale=0.40]{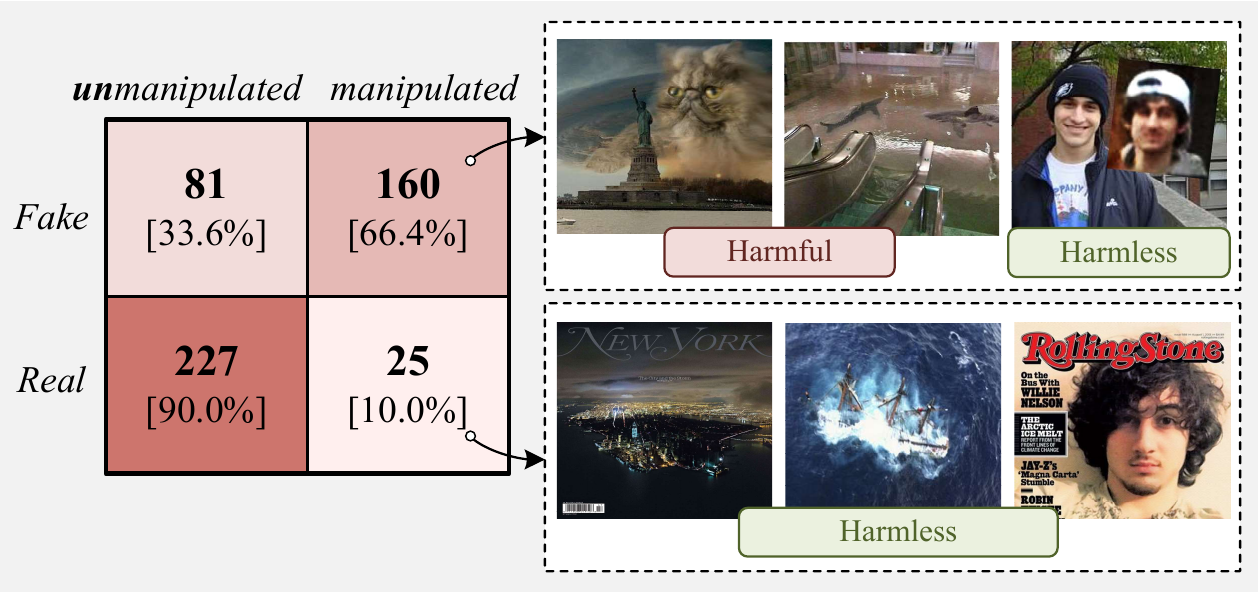}
  \caption{The statistics from the MMD dataset \textit{Twitter} reveal a quantitative relationship between manipulated visual content and veracity labels. We utilize a pre-trained model for image manipulation detection to determine if an image has been manipulated. Additionally, we provide various examples of images that have been manipulated with both harmful and harmless intentions.}
  \label{example}
\end{figure}

In recent years, widely used social media platforms have connected individuals from all over the world and simplified information sharing. However, as these platforms have grown, misinformation with harmful intentions has also spread extensively, posing risks to individual's mental well-being and financial assets \cite{vosoughi2018spread,van2022misinformation}.
For example, during the fire at Notre Dame Cathedral in Paris in 2019, some conspiracy theorists claimed that the fire was a case of deliberate arson rather than an accident.\footnote{https://abcnews.go.com/International/france-marks-3rd-anniversary-notre-dame-cathedral-fire/story?id=84075948} Although this claim was unverified, it caused extensive panic and distrust among the public.
To mitigate these negative impacts, it is essential to detect misinformation automatically, leading to the emergence of an active research topic known as \textbf{M}isinformation \textbf{D}etection (\textbf{MD}).


Generally, the goal of MD is to develop a veracity predictor that can automatically discriminate the veracity label of an article, \eg real and fake. Previous MD efforts have focused on encoding raw articles into a high-dimensional semantic space and learning potential relationships between these semantics and their veracity labels using various deep models \cite{zhang2021mining,sheng2022zoom,zhu2022generalizing}. However, most existing MD approaches only deal with text-only articles, which is not realistic given the prevalence of multimodal content on social media platforms today.  To address this, recent efforts have been directed towards developing \textbf{M}ultimodal \textbf{M}isinformation \textbf{D}etection (\textbf{MMD}) approaches to meet this practical need, which can detect misinformation across multiple modalities, \eg text, image, and video. 
The typical MMD pipelines first extract unimodal semantic features using various prevalent feature extractors \cite{he2016deep,devlin2019bert}. These features are then aligned and integrated into a multimodal feature to predict the veracity labels \cite{qian2021hierarchical,wang2023cross,chen2022cross}.
Upon this pipeline, state-of-the-art MMD approaches develop innovative multimodal interaction strategies to fuse semantic features \cite{ying2023bootstrapping,jin2017multimodal,qian2021hierarchical}, and model the semantic inconsistencies between different modalities \cite{chen2022cross,qi2021improving,sun2021inconsistency}.

While modality features can enhance current MMD methods, these approaches often treat MMD as a conventional text classification problem that relies heavily on sample semantics. However, misinformation is a nuanced issue, with its veracity influenced by multiple factors.
As explored in prior studies \cite{cao2020exploring,bu2023combating}, many fake articles tend to include manipulated visual elements generated through various techniques, such as image copy-move manipulation and video editing \cite{cozzolino2015efficient}. To investigate this perspective, we perform an initial statistical analysis on a public MMD dataset, \textit{Twitter}, illustrated in Fig.~\ref{example}. The results present two key findings.
Initially, we observe that about 66.4\% of fake articles include manipulated visual content, suggesting that such content could serve as a useful indicator for identifying fake articles. Conversely, we find that about 10.0\% of real articles also contain manipulated visual elements, which seems to contradict the general assumption that real articles should be entirely authentic. Upon further examination of these manipulated visuals, we empirically observe that those in fake articles tend to be associated with harmful intentions, such as deception or pranks, while the manipulated content in real articles is typically driven by harmless intentions, like watermarking or aesthetic modifications, as illustrated in Fig.~\ref{example}. Based on these observations and the intention-based perspective \cite{da2021edited}, we hypothesize that visual content manipulated with harmful intent can serve as a strong indicator for detecting misinformation.
Driven by these insights, we propose a method for detecting misinformation by extracting distinctive \textit{manipulation features} that indicate whether the visual content has been manipulated, alongside \textit{intention features} that distinguish between harmful and harmless intentions behind the manipulation. To this end, we introduce a novel MMD framework, named \textbf{HA}rmful \textbf{Vi}sual \textbf{C}ontent \textbf{M}anipulation \textbf{M}atters in \textbf{MMD} (\textbf{\baby}).
In particular, we extract manipulation and intention features from multimodal articles and leverage them to define two binary classification tasks: manipulation classification and intention classification. These classifiers are then trained using their respective binary labels. However, the ground-truth labels for these tasks are \textbf{unknown} in prevalent MMD datasets. To overcome this challenge, we propose two weakly supervised signals as alternatives to these labels.
To supervise the manipulation classifier, we adopt a knowledge distillation approach \cite{hinton2015distilling,zhou2023bridging} to train a manipulation teacher capable of identifying whether the visual content has been manipulated. The teacher’s discriminative abilities are then transferred to the manipulation classifier. Specifically, we use additional benchmark datasets for \textbf{I}mage \textbf{M}anipulation \textbf{D}etection (\textbf{IMD}) \cite{dong2013casia} to pre-train the manipulation teacher. For video content, we design a cross-similarity module that refines the pseudo labels generated by the teacher. To address the distribution shift between IMD and MMD data, we generate synthetic manipulated images from MMD datasets and introduce a \textbf{P}ositive and \textbf{U}nlabeled (\textbf{PU}) learning objective to adapt the teacher to MMD data.
Second, considering a fact that \textit{if the visual content in real information has been manipulated, the manipulation is likely to have a harmless intention}, we can treat the intention classification task as a PU learning problem and address it using a PU approach.

We validate the performance of the method \baby across 4 prevalent MMD datasets, including \textit{GossipCop} \cite{shu2020fakenewsnet}, \textit{Weibo} \cite{jin2017multimodal}, \textit{Twitter} \cite{boididou2018detection}, and \textit{FakeSV} \cite{qi2023fakesv}, and compare its performance against several baseline MMD models. The results show a consistent enhancement in average performance across all metrics when using \baby, highlighting its effectiveness.

In summary, our contributions are the following three-fold:
\begin{itemize}
    \item We propose that the manipulation of visual content, along with its underlying intention, plays a significant role in MMD. To capture and integrate these manipulation and intention features, we introduce a novel MMD model, \baby.
    \item To address the challenge of unknown manipulation and intention labels, we design two weakly-supervised cues using supplemental IMD datasets and the PU learning technique.
    \item Comprehensive experiments are carried out on four MMD datasets, showcasing the performance enhancements achieved by \baby.
\end{itemize}

\noindent \textbf{Remark.} 
This paper is an extension version of our previous conference paper \cite{wang2024harmfully}. This paper extends our earlier work in several important aspects: 

(1) We generalize our previous method by designing a new video manipulation detection method to correct the pseudo labels generated by the manipulation teacher (Sec.~\ref{sec:method});

(2) We evaluate \baby across four benchmark MMD dataset, demonstrating its effectiveness (Sec.~\ref{sec:experiment}).

\section{Related Work}

In this section, we briefly describe the related literature on misinformation detection, manipulation detection, and positive and unlabeled learning.

\subsection{Misinformation Detection}

Recent MD models can be categorized into text-only and multimodal methods.

\noindent \textbf{Text-only MD.}
Text-only models typically formulate MD as a binary classification task, aiming to learn the potential correlation between textual features and veracity labels. Previous research has predominantly focused on augmenting these models with additional discriminative signals, such as world knowledge \cite{dun2021kan,zhou2024finefake}, the intentions behind news creators \cite{wang2024why}, domain-specific information \cite{nan2021mdfend,zhu2023memory}, and stylistic attributes of writing \cite{zhu2023memory}. In parallel, with the rapid advancements in \textbf{L}arge \textbf{L}anguage \textbf{M}odels (\textbf{LLMs}), the field has seen a growing interest in leveraging their capabilities to support MD. These approaches include prompting LLMs to generate explanatory rationales for detecting potential misinformation in news articles \cite{hu2024bad} or employing LLMs to retrieve relevant evidence from large-scale knowledge bases to enhance detection accuracy \cite{yue2024evidence,li2024re}.


\noindent \textbf{Multimodal MD.}
Currently, most MMD models predominantly addresses the scenario in which an article comprises a single text–image pair. In this setting, misinformation is identified by jointly leveraging the semantic features of both modalities while explicitly modeling their interactions.
Broadly, such cross-modal interactions can be classified into three categories: multimodal alignment, multimodal inconsistency, and multimodal fusion.
Recent arts on multimodal alignment typically employs variational encoder networks \cite{khattar2019mvae,chen2022cross} and contrastive learning techniques \cite{wang2023cross} to align the semantic features of text and image modalities. Ones on multimodal inconsistency is grounded in the assumption that \textit{if the content expressed by images and text is inconsistent, the article is more likely to be fake}. Accordingly, semantic-based \cite{qi2021improving,wei2023modeling}, distribution-based \cite{chen2022cross}, and knowledge-based \cite{sun2021inconsistency,gao2024knowledge} methods have been proposed to capture such cross-modal inconsistencies. Finally, multimodal fusion aims to devise more effective approaches to integrate features of image and text modalities, \eg using attention mechanisms \cite{wu2021multimodal,ying2023bootstrapping} and weighted concatenation methods \cite{wang2024escaping}. 
With the proliferation of short-video platforms, this conventional text–image paradigm can be naturally extended to encompass video, audio, and textual modalities—a formulation referred to as misinformation video detection (MVD). Despite its growing practical relevance, MVD has thus far received limited scholarly attention \cite{qi2023two,bu2023combating}. Existing efforts are primarily directed toward the construction of benchmark datasets and baseline models for this task, such as FakeSV \cite{qi2023fakesv} and FakeTT \cite{bu2024fakingrecipe}.

\subsection{Manipulation Detection}

With the rapid advancement of multimedia technologies, the manipulation of digital media content, \eg images and videos, has become increasingly effortless, in part due to techniques like copy–move \cite{cozzolino2015efficient} and splicing \cite{huh2018fighting}. To automatically control the misuse of these techniques, IMD has become a crucial technique in the community \cite{da2021edited}.
Briefly, IMD seeks to determine whether an image has been altered and to accurately localize the manipulated regions, rendering it a particularly challenging fine-grained segmentation task. Current approaches primarily center on developing powerful neural architectures capable of extracting discriminative semantic features while capturing subtle manipulation artifacts \cite{ma2023iml,sun2023safl}.
For example, Objectformer \cite{wang2022objectformer} leverages a Transformer-based framework \cite{dosovitskiy2021an} to learn patch-level embeddings that model object-level consistencies across different image regions; 
MVSS-Net \cite{chen2021image,dong2023mvss} introduces a multi-view learning paradigm that simultaneously exploits boundary artifacts and learns semantic-agnostic features for robust generalization. 
Additionally, several studies have enriched IMD training resources by synthesizing tampered images from real-world data, thereby expanding the manipulated class and improving detection performance \cite{wu2018busternet,wu2019mantra}.
Motivated by these arts, we introduce a manipulation teacher model, pre-trained on both a benchmark IMD dataset and a synthesized dataset derived from MMD corpora, to provide robust manipulation-aware knowledge for our downstream tasks.

\subsection{Positive and Unlabeled Learning}

PU learning is a prevalent paradigm in which the goal is to develop a binary classifier using only a subset of labeled positive samples and a set of unlabeled instances. Traditional PU learning methods generally fall into two categories: sample-selection and cost-sensitive techniques. Sample-selection approaches apply heuristic strategies to identify likely negative instances within the unlabeled data, which are then used in a supervised or semi-supervised learning framework for classifier training \cite{yu2004pebl,hsieh2019classification,zhang2019boosting}. For instance, PULUS \cite{luo2021pulns} employs reinforcement learning to train a negative sample selector based on the rewards from validation performance.
In contrast, cost-sensitive methods focus on constructing diverse empirical risk functions for negative samples to ensure unbiased risk estimation \cite{plessis2014analysis,kiryo2017positive,chen2020self,li2024positive}. For example, uPU \cite{plessis2014analysis} initially proposed unbiased risk estimation, whereas nnPU \cite{kiryo2017positive} observed that uPU often overfits negative samples, especially in deep learning, and thus introduced a non-negative PU risk by bounding the risk estimation.
Additionally, some recent methods focus on assigning reliable pseudo-labels to unlabeled samples \cite{wang2023beyond} and designing effective data augmentation techniques \cite{li2022who}. 
Within the MMD field, recent studies have also explored PU learning for misinformation detection \cite{souza2022a,wang2024positive}, addressing a weakly-supervised task where detectors are trained using partially labeled real articles as positive samples and treating other articles as unlabeled. Unlike these approaches, our \baby framework leverages PU learning to adapt the pre-trained IMD model to MMD datasets and to enable learning without pre-defined intention labels.

\section{Proposed \baby Method} \label{sec:method}

In this section, we will introduce the definition of our generalized MMD task, and the proposed MMD model \textbf{\baby} in more detail.

\begin{table}[t]
\centering
\renewcommand\arraystretch{1.20}
  \caption{Summary of notations.}
  \label{notation}
  \setlength{\tabcolsep}{5pt}{
  \begin{tabular}{m{2.1cm}<{\centering}|m{5.8cm}<{\centering}}
    \bottomrule
    Notation & Description \\
    \hline
    $\mathbf{x}_i^t$ & text content \\
    $\mathcal{X}_i^v = \{\mathbf{x}_{ij}^v\}_{j=1}^K$ & visual content \\
    $y_i$ & veracity label \\
    $N$ & number of samples \\
    $K$ & number of images \\
    $\boldsymbol{\theta}^t, \boldsymbol{\theta}^v$ & parameters of text and visual encoders \\
    $\boldsymbol{\theta}^m, \boldsymbol{\theta}^e$ & parameters of manipulation and intention encoders \\
    $\boldsymbol{\Pi}$ & parameters of manipulation teacher \\
    $\mathbf{e}_i^t, \mathbf{e}_i^v$ & text and visual features \\
    $\mathbf{e}_i^m, \mathbf{e}_i^e$ & manipulation and intention features \\
    $y^m, y^e$ & pseudo manipulation and intention labels \\
    \bottomrule
  \end{tabular} }
\end{table}

\vspace{2pt} \noindent
\textbf{Problem definition.}
Typically, an MMD dataset consists of $N$ training samples, expressed as $\mathcal{D} = \{(\mathbf{x}_i^t, \mathcal{X}_i^v, y_i)\}_{i=1}^N$, where $\mathbf{x}_i^t$ denotes the text content, $\mathcal{X}_i^v = \{\mathbf{x}_{ij}^v\}_{j=1}^K$ represents the visual content of the $i$-th article, $y_i \in \{0,1\}$ is the corresponding veracity label (0/1 indicates real/fake), and $K$ denotes the number of images or video frames within the visual content. When $K=1$, the task will be degenerated into the typical MMD scenario involving text-image pairs.
The primary goal of the MMD task is to train a misinformation detector capable of predicting the veracity label of any previously unseen article.
The basic pipeline of current MMD approaches typically involves three components: feature encoder, feature fusion network, and predictor. Specifically, the feature encoder extracts the unimodal features from $\mathbf{x}_i^t$ and $\mathcal{X}_i^v$. The feature fusion network then fuses the features into a unified multimodal feature, which is then inputted into the predictor module for veracity classification.
For clarity, the important notations and their descriptions are listed in Table~\ref{notation}.



\begin{figure*}[t]
  \centering
  \includegraphics[scale=0.70]{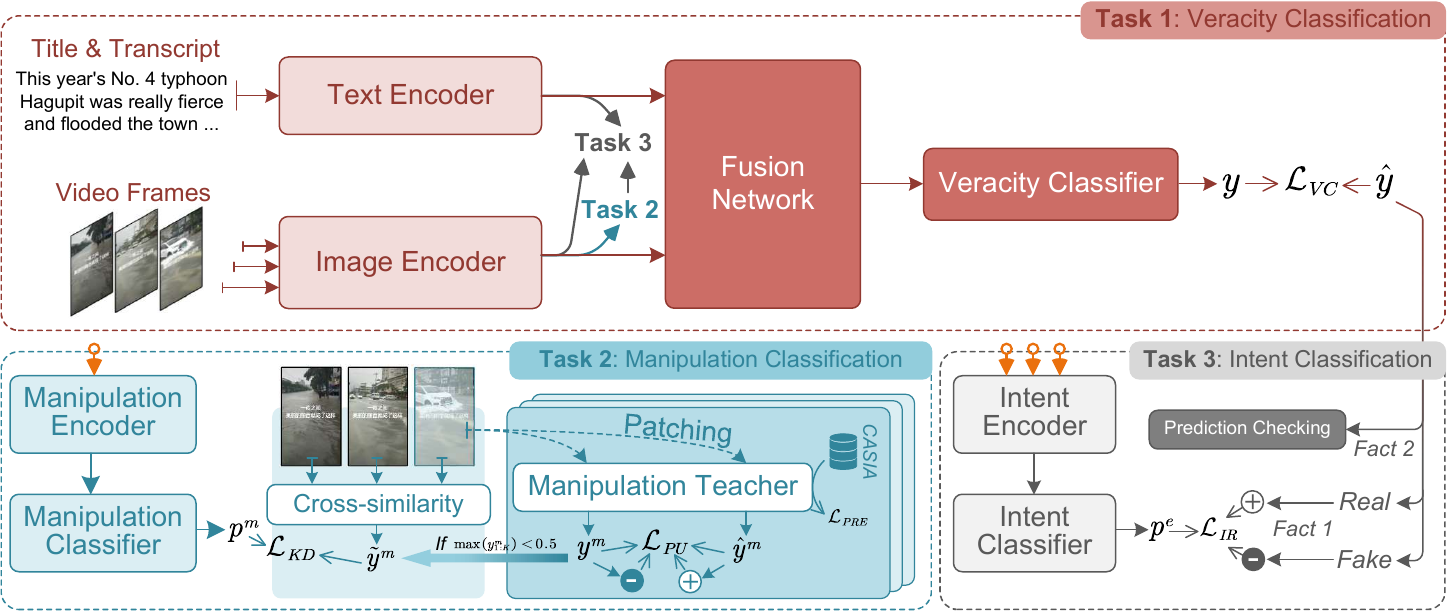}
  \caption{The \baby framework comprises four primary encoders: text encoder, image encoder, manipulation encoder, and intention encoder, which extract features from the provided text content $\mathbf{x}_i^t$ and visual content $\mathcal{X}_i^v$. These features are subsequently fed into a feature fusion network to create a combined representation. To accomplish the multiple tasks, we employ three distinct predictors aimed at veracity classification, manipulation classification, and intention classification.}
  \label{framework}
\end{figure*}

\subsection{Overview of \baby}

We draw motivation from the observation and hypothesis that fake articles often involve visual content that has been manipulated in harmful ways. Thus, for each article, we extract hidden features related to manipulation and harmfulness, which are then integrated with semantic features to produce a more distinctive representation. To approximate these hidden features, we perform manipulation and harmfulness classification as auxiliary tasks. Building on these concepts, we propose \baby within a multi-task learning framework that simultaneously addresses the primary veracity classification task along with the two auxiliary tasks.
More specifically, \baby consists of three main components: a feature encoders module, a feature fusion module, and a predictors module. A comprehensive view of the \baby framework is presented in Fig.~\ref{framework}. The following sections will provide an in-depth explanation of each module.


\vspace{3pt} \noindent
\textbf{Feature encoders module.} 
This module comprises four distinct feature encoding sub-modules: a text encoder, a visual encoder, a manipulation encoder, and an intention encoder. 

For a given text $\mathbf{x}_i^t$ and a set of visual contents $\mathcal{X}_i^v$, the text and visual encoders capture the corresponding text and visual features, denoted as $\mathbf{e}_i^t$ and $\mathbf{e}_i^v$, respectively. To be specific, the text feature $\mathbf{e}_i^t = \mathcal{F}_{\boldsymbol{\theta}^t}(\mathbf{x}_i^t)$ is obtained using a pre-trained BERT model \cite{devlin2019bert} and the visual features $\mathbf{e}_i^v = \mathcal{F}_{\boldsymbol{\theta}^v}(\mathbf{x}_i^v)$ is derived using a ResNet34 model \cite{he2016deep}.
If there are $K \geq 2$ images in $\mathcal{X}_i^v$, we compute the average of their visual features to obtain $\mathbf{e}_i^v$. These features are subsequently projected into a shared feature space via two feed-forward neural networks.
Then, the visual feature $\mathbf{e}_i^v$ is fed into a manipulation encoder to produce its manipulation feature $\mathbf{e}_i^m = \mathcal{F}_{\boldsymbol{\theta}^m}(\mathbf{e}_i^v)$. This manipulation feature is then concatenated with the text and visual features $\mathbf{e}_i^t$ and $\mathbf{e}_i^v$ in an intention encoder, yielding the intention feature $\mathbf{e}_i^e = \mathcal{F}_{\boldsymbol{\theta}^e}(\mathbf{e}_i^t, \mathbf{e}_i^v, \mathbf{e}_i^m)$.


\vspace{3pt} \noindent
\textbf{Feature fusion module.}
With the captured features, the feature fusion module employs an cross-attention mechanism to fuse them into a unified feature $\mathbf{z}_i = \mathcal{F}_{\boldsymbol{\Psi}^f}(\mathbf{e}_i^t, \mathbf{e}_i^v, \mathbf{e}_i^m, \mathbf{e}_i^e)$. The number of heads is fixed to 4 in our experiments.

\vspace{3pt} \noindent
\textbf{Predictors module.}
The module comprises three predictors, each trained on a distinct task: veracity classification, manipulation classification, and intention classification. With the feature $\mathbf{z}_i$, a linear classifier for veracity prediction is utilized, yielding the veracity label prediction as $p_i = \mathbf{W}_V \mathbf{z}_i$. The objective for the veracity classification task over the dataset $\mathcal{D}$ can be expressed as follows:
\begin{equation}
    \label{eq1}
    \mathcal{L}_{VC} = \frac{1}{N} \sum \nolimits _{i=1}^N \ell_{CE} \left(p_i , y_i \right),
\end{equation}
where $\ell_{CE}(\cdot, \cdot)$ represents the cross-entropy loss function. Using the manipulation feature $\mathbf{e}_i^m$ and intention feature $\mathbf{e}_i^e$, we apply their respective classifiers to obtain predictions $p_i^m = \mathbf{W}_M \mathbf{e}_i^m \in [0, 1]$ and $p_i^e = \mathbf{W}_E \mathbf{e}_i^e \in [0, 1]$, where $p_i^m = 1$ or $0$ indicates whether the visual content has undergone manipulation, while $p_i^e = 1$ or $0$ represents whether the manipulation is performed with harmless or harmful intention, respectively.

Unfortunately, in MMD datasets, the ground-truth manipulation and intention labels remain \textbf{unknown}. To circumvent this challenge, we propose two weakly supervised cues as alternatives for the manipulation and intention classification tasks.
For the manipulation classification task, we first train a teacher model $f_{\boldsymbol{\Pi}}(\cdot)$ on auxiliary IMD datasets, such as \textit{CASIAv2} \cite{dong2013casia}, by utilizing a loss function $\mathcal{L}_{PRE}$. To mitigate the data distribution discrepancy problem between the IMD and MMD datasets, we further adapt the teacher model using a PU loss $\mathcal{L}_{PU}$. With the prediction $y_i^m = f_{\boldsymbol{\Pi}}(\mathbf{x}_i^v)$ from the teacher model, we employ a cross-similarity module to refine it and distill its knowledge to the prediction $p_i^m$ as follows:
\begin{equation}
    \label{eq2}
    \mathcal{L}_{KD} = \frac{1}{N} \sum \nolimits_{i=1}^N D_{KL} \big( y_i^m, p_i^m \big),
\end{equation}
where $D_{KL}(\cdot\ , \cdot)$ represents the KL divergence function. For the intention classification task, we are guided by the fact that \textit{if the visual content of the \textbf{real} article is manipulated, its intention must be \textbf{harmless}}, leading us to reformulate the intention classification as a PU learning problem, defined by the objective $\mathcal{L}_{IR}$. Furthermore, considering the fact that \textit{if the visual content of one article is manipulated with a \textbf{harmful} intention, the veracity label of this article must be \textbf{fake}}, we can also assess the dependability of the prediction $p_i^e$ and exclude unreliable examples during the training process.

Building upon these tasks, our primary objectives are outlined as follows:
\begin{equation}
    \label{eq3}
    \mathop{\boldsymbol{\min}} \limits _{\boldsymbol{\theta}} \mathcal{L} = \mathcal{L}_{VC} + \alpha \mathcal{L}_{KD} + \beta \mathcal{L}_{IR},
\end{equation}
\begin{equation}
    \label{eq4}
     \mathop{\boldsymbol{\min}} \limits _{\boldsymbol{\Pi}} \mathcal{L}_{\text{teacher}} = \mathcal{L}_{PRE} + \delta \mathcal{L}_{PU},
\end{equation}
where $\alpha$, $\beta$, and $\delta$ serve as hyperparameters that mediate the equilibrium between various loss functions.
We will alternatively optimize the detector and the teacher model parameterized by $\boldsymbol{\theta}$ and $\boldsymbol{\Pi}$ by the objectives in Eqs.~\eqref{eq3} and \eqref{eq4}.
For clarity, the complete training process is detailed in Alg.~\ref{algorithm}. Subsequent sections will provide an in-depth description of the manipulation and intention classification tasks.

\subsection{Manipulation Classification}

Typically, manipulation classification entails first training a manipulation teacher model, denoted as $f_{\boldsymbol{\Pi}}(\cdot)$, followed by distilling its predictions into the output $q_i^m$ as defined in Eq.~\eqref{eq2}. The teacher model’s optimization focuses on two primary objectives: a pre-training objective, $\mathcal{L}_{PRE}$, and an adaptation objective, $\mathcal{L}_{PU}$.

We introduce a prevalent IMD dataset, termed $\mathcal{D}_{\mu} = \{\mathbf{x}_i^{\mu}, y_i^{\mu}\}_{i=1}^{N_\mu}$, such as \textit{CASIAv2} \cite{dong2013casia}, which includes $N_\mu$ images $\mathbf{x}^{\mu}$ paired with their corresponding manipulation labels $y^{\mu} \in \{0, 1\}$, where $y^{\mu} = 1$ or $0$ indicates whether $\mathbf{x}^{\mu}$ is manipulated. Each image $\mathbf{x}_i^{\mu}$ is then fed into a ResNet18 model, which underpins our teacher model.
Recent IMD research highlights that identifying manipulation cues requires both semantic information and subtle image details \cite{sun2023safl, dong2023mvss}. In line with this, we utilize the method from \cite{chen2021image} to extract features from intermediate layers of ResNet18, integrate them using a self-attention mechanism, and predict the manipulation label for $\mathbf{x}_i^{\mu}$. The associated objective is represented as follows:
\begin{equation}
    \label{eq5}
    \mathcal{L}_{PRE} = \frac{1}{N_\mu} \sum \nolimits _{i=1}^{N_\mu} \ell_{CE} \left( f_{\boldsymbol{\Pi}} \big( \mathbf{x}_i^{\mu} \big), y_i^{\mu} \right).
\end{equation}

Typically, the teacher model for manipulation detection in our method is a plug-and-play model. As the research community advances, this component can be replaced with more advanced image manipulation detection models.
Then, confronting the unavoidable data distribution discrepancy issue between the IMD dataset $\mathcal{D}_{\mu}$ and the MMD dataset $\mathcal{D}$, we suggest adapting the teacher, pre-trained on $\mathcal{D}_{\mu}$, to the MMD dataset via a PU learning scheme. To be specific, for an image or a video frame $\mathbf{x}_{ij}^v$ drawn from $\mathcal{X}_i^v$ in $\mathcal{D}$, we manipulate it using the image copy-moving technique \cite{cozzolino2015efficient}, creating its manipulated counterpart $\mathbf{\hat x}_{ij}^v$. 
Accordingly, the ground-truth manipulation label for $\mathbf{\hat x}_{ij}^v$ is inherently designated as ``\textit{manipulated}'', represented by $\hat y_{ij}^m = 1$, and construct a training subset $\mathcal{P}^m = \{\mathbf{\hat x}_{ij}^v, \hat y_{ij}^m = 1\}_{i=1}^N$. 
Simultaneously, the manipulation label for $\mathbf{x}_{ij}^v$ remains unknown, leading us to form an additional unlabeled subset $\mathcal{U}^m = \{\mathbf{x}_{ij}^v\}_{i=1}^N$. 
Inspired by the PU learning regime, which focuses on learning a binary classifier using a part of labeled positive examples and an abundance of unlabeled examples, we reformulate the manipulation classification over $\mathcal{P}^m \cup \mathcal{U}^m$ as a PU learning problem. 

Formally, in PU learning, various methodologies have been proposed for risk estimation. In our study, we adopt a variational PU learning framework \cite{chen2020a}.\footnote{This variational approach is grounded in the ``\textit{selected completely at random}'' assumption that does not require additional class priors, which aligns with the context of our method. Furthermore, it has also empirically demonstrated notable efficacy in \baby.} Considering two subsets $\mathcal{P}^m \sim \mathbb{P}_P \triangleq \mathbb{P}(\mathbf{x}^v | y^m = 1)$ and $\mathcal{U}^m \sim \mathbb{P}_U \triangleq \mathbb{P}(\mathbf{x}^v)$,\footnote{Given that $\mathbb{P}(\mathbf{\hat x}^v)$ and $\mathbb{P}(\mathbf{x}^v)$ are independently and identically distributed (IID), to keep our notations clear, we uniformly utilize $\mathbf{x}^v$ and $y^m$ to indicate the visual content and its manipulation label.} we use the Bayes rule to estimate the distribution $\mathbb{P}_P$ as follows:
\begin{align}
    \label{eq6}
    \mathbb{P}_P = \frac{\mathbb{P}(y^m=1 | \mathbf{x}^v) \mathbb{P}(\mathbf{x}^v)}{\int \mathbb{P}(y^m=1 | \mathbf{x}^v) \mathbb{P}(\mathbf{x}^v) d \mathbf{x}^v}
    & = \frac{f_{\boldsymbol{\Pi}^\star}(\mathbf{x}^v) \mathbb{P}_U}{\mathbb{E}_{\mathbb{P}_U} \left[ f_{\boldsymbol{\Pi}^\star}(\mathbf{x}^v) \right] } \nonumber \\
    \approx & \frac{f_{\boldsymbol{\Pi}}(\mathbf{x}^v) \mathbb{P}_U}{\mathbb{E}_{\mathbb{P}_U} \left[ f_{\boldsymbol{\Pi}}(\mathbf{x}^v) \right] }
    \triangleq \mathbb{P}_{\boldsymbol{\Pi}},
\end{align}
where $\mathbb{P}_{\boldsymbol{\Pi}}$ denotes the data distribution generated by the teacher model parameterized by ${\boldsymbol{\Pi}}$, and $f_{\boldsymbol{\Pi}^\star}(\cdot)$ represents an optimal teacher model. Accordingly, to optimize $f_{\boldsymbol{\Pi}}(\cdot)$ towards the optimal $f_{\boldsymbol{\Pi}^\star}(\cdot)$, existing works \cite{chen2020a} have proved that it is effective to minimize the KL divergence between $\mathbb{P}_P$ and $\mathbb{P}_{\boldsymbol{\Pi}}$, which is formalized as follows:
\begin{align}
    \label{eq7}
    D_{KL} \left( \mathbb{P}_P\|\mathbb{P}_{\boldsymbol{\Pi}} \right) & = 
    \mathbb{E}_{\mathbb{P}_P} \left[ \log \frac{\mathbb{P}_P(\mathbf{x}^v)}{\mathbb{P}_{\boldsymbol{\Pi}}(\mathbf{x}^v)} \right] \\
    = & \mathbb{E}_{\mathbb{P}_P} \big[ \log f_{\boldsymbol{\Pi}^\star}(\mathbf{x}^v) \big] + \mathbb{E}_{\mathbb{P}_P} \big[ \log \mathbb{P}_P(\mathbf{x}^v) \big] \nonumber \\
    & - \log \mathbb{E}_{\mathbb{P}_U} \big[ f_{\boldsymbol{\Pi}^\star}(\mathbf{x}^v) \big] 
     - \Big( \mathbb{E}_{\mathbb{P}_P} \big[ \log f_{\boldsymbol{\Pi}}(\mathbf{x}^v) \big] \nonumber \\
    & + \mathbb{E}_{\mathbb{P}_P} \big[ \log \mathbb{P}_P(\mathbf{x}^v) \big] 
    - \log \mathbb{E}_{\mathbb{P}_U} \big[ f_{\boldsymbol{\Pi}}(\mathbf{x}^v) \big] \Big). \nonumber
\end{align}
Accordingly, the PU optimization objective is specified as follows:
\begin{equation}
    \label{eq8}
    \mathcal{L}_{PU} \triangleq \log \mathbb{E}_{\mathcal{U}^m \sim \mathbb{P}_U} \big[ f_{\boldsymbol{\Pi}}(\mathbf{x}^v) \big]
    - \mathbb{E}_{\mathcal{P}^m \sim \mathbb{P}_P} \big[ \log f_{\boldsymbol{\Pi}}(\mathbf{x}^v) \big].
\end{equation}

By optimizing $\mathcal{L}_{PRE}$ and $\mathcal{L}_{PU}$ with Eq.~\eqref{eq4}, we can obtain a strong manipulation teacher, which generates the pseudo manipulation label $y_{ij}^m$. Notably, when training with PU learning on the MMD data, the size of the IMD data is significantly larger than that of the MMD data. Therefore, for the objective $\mathcal{L}_{PRE}$, we use an online learning regime to optimize this objective function. Specifically, during multiple training epochs on the MMD data, we consistently sample new data from the IMD data, ensuring that the training IMD data does not overlap.
Additionally, video editing is also a visual content manipulation technique that should be considered. Therefore, we present a cross-similarity method to generate the video manipulation label $\tilde y_i^m$ and use this approach to refine the pseudo labels $y^m$.
Specifically, given the features $\{\mathbf{e}_{ij}^v\}_{j=1}^K$ of $K$ images in $\mathcal{X}_i^v$, we calculate the cosine similarities between different features as $\tilde y_i^m$ as follows:
\begin{equation}
    \label{eq8-1}
    \tilde y_i^m = \nu \left( \frac{1}{K^2} \sum \nolimits _{j=1}^K \sum \nolimits _{k=1}^K \nu \left( {|k-j|}^{-1} \right) \text{cos} \left( \mathbf{e}_{ij}^v, \mathbf{e}_{ik}^v \right) \right),
\end{equation}
where $\nu(\cdot)$ represents a sigmoid activation function. Notably, we design a weight $\nu \big( {|k-j|}^{-1} \big)$, so that the semantic inconsistency between two video frames is considered more significant in measuring video editing if their temporal distance is closer. Based on $\tilde y_i^m$, the refined manipulation label is as follows:
\begin{equation}
    \label{eq8-2}
    y_i^m = \left\{
	\begin{aligned}
	       \  \max \left( \tilde y_i^m, \max \big( y_{i[1:K]}^m \big) \right), & \max \big( y_{i[1:K]}^m \big) < 0.5, \\
	       \  \min \left( \max \big( y_{i[1:K]}^m \big) + \tilde y_i^m, 1 \right), & \max \big( y_{i[1:K]}^m \big) \geq 0.5, \\
	\end{aligned}
	\right .
\end{equation}
where $\max \big( y_{i[1:K]}^m \big) < 0.5$ indicates that no images have been manipulated, in which case the label $y_i^m$ depends on whether the video has been edited; if $\max \big( y_{i[1:K]}^m \big) \geq 0.5$ means that the images have already been manipulated, in this scenario, the visual content of the article is certainly manipulated, and the label of video editing $\tilde y_i^m$ will only affect the predicted probability of the pseudo-label.
Accordingly, we can distill the prediction $y_i^m$ from the teacher to $p_i^m$ with Eq.~\eqref{eq2} \cite{hinton2015distilling,zhou2023bridging}. It is important to note that, during the optimization procedure, we initially apply $\mathcal{L}_{PRE}$ for 10 epochs to warm up the teacher model, helping to mitigate the cold start issue when optimizing $\mathcal{L}_{PU}$.

\renewcommand{\algorithmicrequire}{\textbf{Input:}}
\renewcommand{\algorithmicensure}{\textbf{Output:}}
\begin{algorithm}[t]
    \caption{Training summary of \baby.}
    \label{algorithm}
    \begin{algorithmic}[1]
    \Require Training MMD dataset $\mathcal{D}$; IMD dataset $\mathcal{D}_\mu$; hyper-paramaters $\alpha$, $\beta$, and $\delta$; training iterations $I$.
    \Ensure An MMD model parameterized by $\boldsymbol{\theta}$; teacher model parameterized by $\boldsymbol{\Pi}$.
    \State Initialize $\boldsymbol{\theta}^t$ and $\boldsymbol{\theta}^v$ with their pre-trained weights, and other parameters from scratch.
    \State Warm-up $\boldsymbol{\Pi}$ with $\mathcal{L}_{PRE}$ for 10 epochs.
    \For{$i = 1, 2, \cdots, I$}
    \State Draw mini-batches $\mathcal{B}$, $\mathcal{B}_\mu$ from $\mathcal{D}$, $\mathcal{D}_\mu$ randomly.
    \State Manipulate images in $\mathcal{B}$ and form a manipulated $\mathcal{\hat B}$.
    \State Calculate $\mathcal{L}_{PRE}$ with $\mathcal{B}_\mu$ and $\mathcal{L}_{PU}$ with $\mathcal{B} \cup \mathcal{\hat B}$.
    \State Optimize $\boldsymbol{\Pi}$ with Eq.~\eqref{eq4}.
    \State Calculate $\mathcal{L}_{VC}$, $\mathcal{L}_{KD}$, and $\mathcal{L}_{IR}$ with $\mathcal{B}$.
    \State Optimize $\boldsymbol{\theta}$ with Eq.~\eqref{eq3}.
    \EndFor
    \end{algorithmic}
\end{algorithm}

\subsection{Intention Classification}

Considering the intention feature $\mathbf{e}_i^e$, the objective of the intention classification is to enhance its distinctive capacity in identifying the intention underlying the image manipulation. To address the challenge that ground-truth intention labels are consistently unknown, we introduce two heuristic facts as weak-supervised cues to guide the prediction of $p_i^e$. In detail, the first fact is expressed as follows:

\vspace{5pt} \noindent
\textbf{Fact 1.} \textit{If the visual content of the \textbf{real} article is manipulated, its intention must be \textbf{harmless}; But if the visual content of the \textbf{fake} article is manipulated, its intention may be \textbf{harmful} or \textbf{harmless}.} Written as:
\begin{equation}
    y_i^e = \left\{
	\begin{aligned}
	       \  1 ,& \quad y_i = 0 \wedge y_i^m = 1, \\
	       \  0 \ \text{or}\ 1,& \quad y_i = 1 \wedge y_i^m = 1, \\
	\end{aligned}
	\right . \nonumber
\end{equation}
where $y_i^e$ denotes the intention label of the $i$-th sample.
\vspace{3pt}

Given such fact, we can define a subset $\mathcal{D}^e \subset \mathcal{D}$ where each sample meets the condition $y^m = 1$. We then partition $\mathcal{D}^e$ into a positive subset $\mathcal{P}^e$ (where $y = 0$) and an unlabeled subset $\mathcal{U}^e$ (where $y = 1$). Consequently, the task of intention classification over $\mathcal{P}^e \cup \mathcal{U}^e$ can be reframed as a PU learning problem, with its objective, similar to the equation in Eq.~\eqref{eq8}, expressed as:
\begin{equation}
    \label{eq9}
    \mathcal{L}_{IR} \triangleq \log \mathbb{E}_{\mathcal{U}^e \sim \mathbb{P}_U} [ p^e ] - \mathbb{E}_{\mathcal{P}^e \sim \mathbb{P}_P} [ \log p^e ].
\end{equation}

In addition, another fact is presented as:

\vspace{5pt} \noindent
\textbf{Fact 2.} \textit{If the visual content of one article is manipulated by a \textbf{harmful} intention, the veracity label of this article must be \textbf{fake}; But if the visual content of one article is manipulated by a \textbf{harmless} intention, the veracity label of this article may be \textbf{real} or \textbf{fake}.} Written as:
\begin{equation}
    y_i = \left\{
	\begin{aligned}
	       \  1 ,& \quad y_i^e = 0 \wedge y_i^m = 1, \\
	       \  0 \ \text{or}\ 1,& \quad y_i^e = 0 \wedge y_i^m = 1. \\
	\end{aligned}
	\right . \nonumber
\end{equation}

This fact can serve as a criterion for assessing the accuracy of predictions $p_i^m$ and $p_i^e$. For a sample where $p_i^m = 1$ and $p_i^e = 0$, if its ground-truth veracity label $y_i \neq 1$, at least one of its predictions $p_i^m$ and $p_i^e$ is incorrect. Therefore, we remove such incorrect samples during the optimizing process with Eq.~\eqref{eq3}.






\section{Experiments} \label{sec:experiment}

In this section, we conduct extensive experiments and evaluate the performance of \baby by comparing it with existing MMD baselines.

\subsection{Experimental Settings}
\noindent
\textbf{Datasets.}
We perform our experiments on four well-known MMD datasets \textit{GossipCop} \cite{shu2020fakenewsnet}, \textit{Weibo} \cite{jin2017multimodal}, \textit{Twitter} \cite{boididou2018detection}, and \textit{FakeSV} \cite{qi2023fakesv}. Table~\ref{datasetsta} provides the detailed statistics of each dataset.
\begin{itemize}
    \item \textbf{\textit{GossipCop}} \cite{shu2020fakenewsnet} is sourced from a website that fact-checks celebrity news. It comprises 12,840 text-image pairs, where each image is uniquely paired with an article. The articles are typically long-form entertainment news.
    \item \textbf{\textit{Weibo}} \cite{jin2017multimodal} is collected from a Chinese social media platform and includes 9,528 one-to-one image-text pairs. The content is in Chinese and generally consists of short, informal social posts.
    \item \textbf{\textit{Twitter}} \cite{boididou2018detection} is derived from the English social media platform \textit{X.com} and primarily features informal social posts. Unlike \textit{GossipCop} and \textit{Weibo}, the images and text in \textit{Twitter} do not have a one-to-one correspondence; instead, they exhibit more complex one-to-many or many-to-one relationships. Despite containing 13,924 posts, the dataset includes only 514 unique images, which complicates the extraction of meaningful information from the visual data.
    \item \textbf{\textit{FakeSV}} \cite{qi2023fakesv} is a dataset collected from the Chinese short video platform \textit{Douyin}, consisting of 3,654 entries. Each entry includes video frames, articles, subtitles, and audio content, presenting a significant challenge for integrating multimodal features.
\end{itemize}
For these datasets, we follow previous research methods \cite{wang2018eann} to split each dataset into training, validation, and test sets with a ratio of 7:1:2.

\begin{table}[t]
\centering
\renewcommand\arraystretch{1.1}
  \caption{Statistics of three prevalent MMD datasets.}
  \label{datasetsta}
  \setlength{\tabcolsep}{5pt}{
  \begin{tabular}{m{2.1cm}<{\centering}m{1.1cm}<{\centering}m{1.1cm}<{\centering}m{1.3cm}<{\centering}}
    \toprule
    Dataset & \# \textit{Real} & \# \textit{Fake} & \# \textit{Images} \\
    \hline
    \textit{GossipCop} \cite{shu2020fakenewsnet} & 10,259 & 2,581 & 12,840 \\
    \textit{Weibo} \cite{jin2017multimodal} & 4,779 & 4,749 & 9,528 \\
    \textit{Twitter} \cite{boididou2018detection} & 6,026 & 7,898 & 514 \\
    \textit{FakeSV} \cite{qi2023fakesv} & 1,827 & 1,827 & 3,654 \\
    \bottomrule
  \end{tabular} }
\end{table}

\begin{table*}
\centering
\renewcommand\arraystretch{1.15}
  \caption{Experimental results of \baby across three prevalent MD datasets \textit{GossipCop}, \textit{Weibo}, and \textit{Twitter}. The results marked by * are statistically significant compared to its baseline models, satisfying p-value $<$ 0.05.}
  \label{result}
  \setlength{\tabcolsep}{5pt}{
  \begin{tabular}{m{2.1cm}m{1.35cm}<{\centering}m{1.35cm}<{\centering}m{1.35cm}<{\centering}m{1.35cm}<{\centering}m{1.35cm}<{\centering}m{1.35cm}<{\centering}m{1.35cm}<{\centering}m{1.35cm}<{\centering}m{1.10cm}<{\centering}}
    \toprule
    \multirow{2}{*}{\quad \quad Method} & \multirow{2}{*}{Accuracy} & \multirow{2}{*}{Macro F1} & \multicolumn{3}{c}{Real} & \multicolumn{3}{c}{Fake} & \multirow{2}{*}{Avg. $\Delta$}\\
    \cmidrule (r){4-6} \cmidrule (r){7-9}
    & & & Precision & Recall & F1 & Precision & Recall & F1 & \\
    \hline
    \multicolumn{10}{c}{\small \textbf{Dataset}: \textit{GossipCop}} \\
    
    Base model & 87.77{\scriptsize \color{grayv} $\pm$0.56} & 79.51{\scriptsize \color{grayv} $\pm$0.44} & 91.55{\scriptsize \color{grayv} $\pm$0.41} & 93.36{\scriptsize \color{grayv} $\pm$1.20} & 92.37{\scriptsize \color{grayv} $\pm$0.41} & 69.96{\scriptsize \color{grayv} $\pm$1.10} & 63.30{\scriptsize \color{grayv} $\pm$1.46} & 66.92{\scriptsize \color{grayv} $\pm$0.58} & - \\
    \rowcolor{lightgrayv} \quad + \baby & 88.45{\scriptsize \color{grayv} $\pm$0.20}* & 80.32{\scriptsize \color{grayv} $\pm$0.43}* & 91.93{\scriptsize \color{grayv} $\pm$0.14} & 94.08{\scriptsize \color{grayv} $\pm$0.33}* & 92.83{\scriptsize \color{grayv} $\pm$0.12} & 71.99{\scriptsize \color{grayv} $\pm$0.80}* & 64.59{\scriptsize \color{grayv} $\pm$0.74}* & 67.63{\scriptsize \color{grayv} $\pm$0.78}* & \textbf{+0.90} \\
    
    SAFE \cite{zhou2020safe} & 87.78{\scriptsize \color{grayv} $\pm$0.31} & 79.22{\scriptsize \color{grayv} $\pm$0.49} & 91.22{\scriptsize \color{grayv} $\pm$0.30} & 93.34{\scriptsize \color{grayv} $\pm$0.47} & 92.37{\scriptsize \color{grayv} $\pm$0.20} & 70.66{\scriptsize \color{grayv} $\pm$1.32} & 63.12{\scriptsize \color{grayv} $\pm$1.50} & 66.66{\scriptsize \color{grayv} $\pm$0.84} & - \\
    \rowcolor{lightgrayv} \quad + \baby & 88.53{\scriptsize \color{grayv} $\pm$0.24}* & 79.87{\scriptsize \color{grayv} $\pm$0.30}* & 91.90{\scriptsize \color{grayv} $\pm$0.31}* & 94.32{\scriptsize \color{grayv} $\pm$0.54}* & 92.95{\scriptsize \color{grayv} $\pm$0.20}* & 72.19{\scriptsize \color{grayv} $\pm$1.30}* & 64.44{\scriptsize \color{grayv} $\pm$0.73}* & 67.88{\scriptsize \color{grayv} $\pm$0.51}* & \textbf{+0.96} \\
    
    MCAN \cite{wu2021multimodal} & 87.66{\scriptsize \color{grayv} $\pm$0.59} & 78.89{\scriptsize \color{grayv} $\pm$0.34} & 90.89{\scriptsize \color{grayv} $\pm$0.78} & 94.07{\scriptsize \color{grayv} $\pm$1.27} & 92.19{\scriptsize \color{grayv} $\pm$0.46} & 71.01{\scriptsize \color{grayv} $\pm$1.09} & 60.37{\scriptsize \color{grayv} $\pm$1.21} & 65.29{\scriptsize \color{grayv} $\pm$0.87} & - \\
    \rowcolor{lightgrayv} \quad + \baby & 88.27{\scriptsize \color{grayv} $\pm$0.57}* & 79.87{\scriptsize \color{grayv} $\pm$0.36}* & 91.72{\scriptsize \color{grayv} $\pm$0.35}* & 95.13{\scriptsize \color{grayv} $\pm$1.21}* & 93.05{\scriptsize \color{grayv} $\pm$0.41}* & 72.69{\scriptsize \color{grayv} $\pm$0.96}* & 62.64{\scriptsize \color{grayv} $\pm$1.21}* & 66.65{\scriptsize \color{grayv} $\pm$0.32}* & \textbf{+1.21} \\
    
    CAFE \cite{chen2022cross} & 87.40{\scriptsize \color{grayv} $\pm$0.71} & 79.51{\scriptsize \color{grayv} $\pm$0.61} & 91.07{\scriptsize \color{grayv} $\pm$0.25} & 93.84{\scriptsize \color{grayv} $\pm$1.28} & 92.16{\scriptsize \color{grayv} $\pm$0.50} & 71.60{\scriptsize \color{grayv} $\pm$1.39} & 61.16{\scriptsize \color{grayv} $\pm$1.10} & 66.24{\scriptsize \color{grayv} $\pm$0.72} & - \\
    \rowcolor{lightgrayv} \quad + \baby & 88.18{\scriptsize \color{grayv} $\pm$0.44}* & 80.43{\scriptsize \color{grayv} $\pm$0.48}* & 91.50{\scriptsize \color{grayv} $\pm$0.45} & 94.46{\scriptsize \color{grayv} $\pm$1.00}* & 92.80{\scriptsize \color{grayv} $\pm$0.31}* & 72.84{\scriptsize \color{grayv} $\pm$0.83}* & 62.51{\scriptsize \color{grayv} $\pm$0.90}* & 67.58{\scriptsize \color{grayv} $\pm$0.83}* & \textbf{+0.91} \\
    
    BMR \cite{ying2023bootstrapping} & 87.26{\scriptsize \color{grayv} $\pm$0.46} & 79.03{\scriptsize \color{grayv} $\pm$0.64} & 90.89{\scriptsize \color{grayv} $\pm$0.24} & 93.99{\scriptsize \color{grayv} $\pm$0.59} & 92.14{\scriptsize \color{grayv} $\pm$0.29} & 71.15{\scriptsize \color{grayv} $\pm$1.23} & 60.37{\scriptsize \color{grayv} $\pm$1.21} & 65.51{\scriptsize \color{grayv} $\pm$1.01} & - \\
    \rowcolor{lightgrayv} \quad + \baby & 87.95{\scriptsize \color{grayv} $\pm$0.27}* & 79.99{\scriptsize \color{grayv} $\pm$0.57}* & 91.40{\scriptsize \color{grayv} $\pm$0.51}* & 94.73{\scriptsize \color{grayv} $\pm$0.75}* & 93.14{\scriptsize \color{grayv} $\pm$0.19}* & 72.26{\scriptsize \color{grayv} $\pm$0.73}* & 62.94{\scriptsize \color{grayv} $\pm$0.89}* & 66.80{\scriptsize \color{grayv} $\pm$1.09}* & \textbf{+1.11} \\
    
    \hline
    \specialrule{0em}{0.5pt}{0.5pt}
    \hline
    \multicolumn{10}{c}{\small \textbf{Dataset}: \textit{Weibo}} \\
    Base model & 90.87{\scriptsize \color{grayv} $\pm$0.34} & 90.75{\scriptsize \color{grayv} $\pm$0.34} & 91.08{\scriptsize \color{grayv} $\pm$0.23} & 90.17{\scriptsize \color{grayv} $\pm$0.85} & 90.62{\scriptsize \color{grayv} $\pm$0.40} & 90.87{\scriptsize \color{grayv} $\pm$0.70} & 91.41{\scriptsize \color{grayv} $\pm$0.28} & 91.29{\scriptsize \color{grayv} $\pm$0.29} & - \\
    \rowcolor{lightgrayv} \quad + \baby & 91.62{\scriptsize \color{grayv} $\pm$0.66}* & 91.61{\scriptsize \color{grayv} $\pm$0.66}* & 91.83{\scriptsize \color{grayv} $\pm$0.87}* & 93.23{\scriptsize \color{grayv} $\pm$0.56}* & 91.39{\scriptsize \color{grayv} $\pm$0.76}* & 92.52{\scriptsize \color{grayv} $\pm$0.89}* & 91.87{\scriptsize \color{grayv} $\pm$0.64} & 91.84{\scriptsize \color{grayv} $\pm$0.62}* & \textbf{+1.11} \\
    
    SAFE \cite{zhou2020safe} & 91.06{\scriptsize \color{grayv} $\pm$0.88} & 91.04{\scriptsize \color{grayv} $\pm$0.89} & 91.09{\scriptsize \color{grayv} $\pm$1.25} & 90.51{\scriptsize \color{grayv} $\pm$0.90} & 90.73{\scriptsize \color{grayv} $\pm$1.04} & 91.27{\scriptsize \color{grayv} $\pm$0.78} & 91.57{\scriptsize \color{grayv} $\pm$1.14} & 91.36{\scriptsize \color{grayv} $\pm$0.85} & - \\
    \rowcolor{lightgrayv} \quad + \baby & 92.22{\scriptsize \color{grayv} $\pm$0.91}* & 92.22{\scriptsize \color{grayv} $\pm$0.93}* & 91.15{\scriptsize \color{grayv} $\pm$1.08} & 94.22{\scriptsize \color{grayv} $\pm$0.84}* & 92.14{\scriptsize \color{grayv} $\pm$0.92}* & 94.34{\scriptsize \color{grayv} $\pm$1.00}* & 91.34{\scriptsize \color{grayv} $\pm$1.09} & 92.30{\scriptsize \color{grayv} $\pm$0.66}* & \textbf{+1.42} \\

    MCAN \cite{wu2021multimodal} & 90.99{\scriptsize \color{grayv} $\pm$0.83} & 90.99{\scriptsize \color{grayv} $\pm$0.83} & 89.66{\scriptsize \color{grayv} $\pm$0.82} & 92.24{\scriptsize \color{grayv} $\pm$1.10} & 90.81{\scriptsize \color{grayv} $\pm$0.90} & 92.69{\scriptsize \color{grayv} $\pm$0.80} & 89.92{\scriptsize \color{grayv} $\pm$0.99} & 91.20{\scriptsize \color{grayv} $\pm$0.79} & - \\
    \rowcolor{lightgrayv} \quad + \baby & 92.01{\scriptsize \color{grayv} $\pm$0.80}* & 92.01{\scriptsize \color{grayv} $\pm$0.80}* & 90.44{\scriptsize \color{grayv} $\pm$0.70}* & 93.37{\scriptsize \color{grayv} $\pm$0.87}* & 91.88{\scriptsize \color{grayv} $\pm$0.85}* & 93.59{\scriptsize \color{grayv} $\pm$0.74}* & 90.84{\scriptsize \color{grayv} $\pm$0.78}* & 92.17{\scriptsize \color{grayv} $\pm$0.76}* & \textbf{+0.98} \\

    CAFE \cite{chen2022cross} & 90.99{\scriptsize \color{grayv} $\pm$0.78} & 90.98{\scriptsize \color{grayv} $\pm$0.78} & 90.31{\scriptsize \color{grayv} $\pm$0.72} & 91.19{\scriptsize \color{grayv} $\pm$1.09} & 90.73{\scriptsize \color{grayv} $\pm$0.97} & 91.70{\scriptsize \color{grayv} $\pm$1.26} & 90.81{\scriptsize \color{grayv} $\pm$1.03} & 91.24{\scriptsize \color{grayv} $\pm$0.60} & - \\
    \rowcolor{lightgrayv} \quad + \baby & 91.95{\scriptsize \color{grayv} $\pm$1.06}* & 91.84{\scriptsize \color{grayv} $\pm$1.01}* & 91.25{\scriptsize \color{grayv} $\pm$0.55}* & 92.38{\scriptsize \color{grayv} $\pm$1.04}* & 91.66{\scriptsize \color{grayv} $\pm$0.91}* & 92.99{\scriptsize \color{grayv} $\pm$0.83}* & 91.93{\scriptsize \color{grayv} $\pm$0.91}* & 92.11{\scriptsize \color{grayv} $\pm$0.75}* & \textbf{+1.02} \\

    BMR \cite{ying2023bootstrapping} & 90.17{\scriptsize \color{grayv} $\pm$0.92} & 90.15{\scriptsize \color{grayv} $\pm$0.93} & 90.09{\scriptsize \color{grayv} $\pm$1.20} & 89.60{\scriptsize \color{grayv} $\pm$0.85} & 89.81{\scriptsize \color{grayv} $\pm$1.00} & 90.36{\scriptsize \color{grayv} $\pm$0.93} & 90.71{\scriptsize \color{grayv} $\pm$0.78} & 90.50{\scriptsize \color{grayv} $\pm$0.81} & - \\
    \rowcolor{lightgrayv} \quad + \baby & 91.74{\scriptsize \color{grayv} $\pm$0.40}* & 91.68{\scriptsize \color{grayv} $\pm$0.40}* & 91.01{\scriptsize \color{grayv} $\pm$0.92}* & 93.17{\scriptsize \color{grayv} $\pm$0.82}* & 91.56{\scriptsize \color{grayv} $\pm$0.43}* & 93.40{\scriptsize \color{grayv} $\pm$0.84}* & 91.29{\scriptsize \color{grayv} $\pm$0.67}* & 91.81{\scriptsize \color{grayv} $\pm$0.38}* & \textbf{+1.79} \\
    
    \hline
    \specialrule{0em}{0.5pt}{0.5pt}
    \hline
    \multicolumn{10}{c}{\small \textbf{Dataset}: \textit{Twitter}} \\
    Base model & 65.08{\scriptsize \color{grayv} $\pm$1.18} & 63.91{\scriptsize \color{grayv} $\pm$1.09} & 57.29{\scriptsize \color{grayv} $\pm$1.26} & 66.67{\scriptsize \color{grayv} $\pm$1.01} & 61.48{\scriptsize \color{grayv} $\pm$1.56} & 72.04{\scriptsize \color{grayv} $\pm$0.96} & 62.41{\scriptsize \color{grayv} $\pm$0.92} & 65.35{\scriptsize \color{grayv} $\pm$1.01} & - \\
    \rowcolor{lightgrayv} \quad + \baby & 66.27{\scriptsize \color{grayv} $\pm$0.66}* & 65.67{\scriptsize \color{grayv} $\pm$1.27}* & 59.70{\scriptsize \color{grayv} $\pm$1.16}* & 69.70{\scriptsize \color{grayv} $\pm$0.71}* & 62.46{\scriptsize \color{grayv} $\pm$1.08}* & 73.19{\scriptsize \color{grayv} $\pm$0.93}* & 64.12{\scriptsize \color{grayv} $\pm$1.12}* & 67.86{\scriptsize \color{grayv} $\pm$0.82}* & \textbf{+1.84} \\

    SAFE \cite{zhou2020safe} & 66.43{\scriptsize \color{grayv} $\pm$0.33} & 66.33{\scriptsize \color{grayv} $\pm$0.32} & 58.28{\scriptsize \color{grayv} $\pm$0.50} & 73.63{\scriptsize \color{grayv} $\pm$1.38} & 64.47{\scriptsize \color{grayv} $\pm$0.53} & 74.94{\scriptsize \color{grayv} $\pm$0.84} & 61.78{\scriptsize \color{grayv} $\pm$1.26} & 68.34{\scriptsize \color{grayv} $\pm$0.69} & - \\
    \rowcolor{lightgrayv} \quad + \baby & 67.15{\scriptsize \color{grayv} $\pm$0.96}* & 67.00{\scriptsize \color{grayv} $\pm$0.89}* & 59.32{\scriptsize \color{grayv} $\pm$0.90}* & 74.05{\scriptsize \color{grayv} $\pm$0.99} & 65.65{\scriptsize \color{grayv} $\pm$0.70}* & 76.49{\scriptsize \color{grayv} $\pm$0.60}* & 63.58{\scriptsize \color{grayv} $\pm$1.09}* & 68.77{\scriptsize \color{grayv} $\pm$0.94} & \textbf{+0.98} \\

    MCAN \cite{wu2021multimodal} & 65.82{\scriptsize \color{grayv} $\pm$0.64} & 65.24{\scriptsize \color{grayv} $\pm$1.34} & 58.30{\scriptsize \color{grayv} $\pm$1.07} & 63.66{\scriptsize \color{grayv} $\pm$1.03} & 61.16{\scriptsize \color{grayv} $\pm$1.23} & 71.70{\scriptsize \color{grayv} $\pm$1.03} & 67.42{\scriptsize \color{grayv} $\pm$1.39} & 69.33{\scriptsize \color{grayv} $\pm$1.22} & - \\
    \rowcolor{lightgrayv} \quad + \baby & 67.14{\scriptsize \color{grayv} $\pm$1.11}* & 66.58{\scriptsize \color{grayv} $\pm$1.21}* & 60.63{\scriptsize \color{grayv} $\pm$0.99}* & 64.94{\scriptsize \color{grayv} $\pm$1.04}* & 62.55{\scriptsize \color{grayv} $\pm$1.28}* & 72.86{\scriptsize \color{grayv} $\pm$0.82}* & 68.77{\scriptsize \color{grayv} $\pm$1.12}* & 70.61{\scriptsize \color{grayv} $\pm$1.10}* & \textbf{+1.43} \\

    CAFE \cite{chen2022cross} & 65.62{\scriptsize \color{grayv} $\pm$0.58} & 65.04{\scriptsize \color{grayv} $\pm$0.48} & 58.39{\scriptsize \color{grayv} $\pm$0.90} & 66.24{\scriptsize \color{grayv} $\pm$1.48} & 62.05{\scriptsize \color{grayv} $\pm$0.21} & 72.37{\scriptsize \color{grayv} $\pm$1.28} & 65.16{\scriptsize \color{grayv} $\pm$1.06} & 68.57{\scriptsize \color{grayv} $\pm$1.05} & - \\
    \rowcolor{lightgrayv} \quad + \baby & 65.89{\scriptsize \color{grayv} $\pm$1.30} & 65.37{\scriptsize \color{grayv} $\pm$0.87} & 59.91{\scriptsize \color{grayv} $\pm$0.55}* & 67.28{\scriptsize \color{grayv} $\pm$1.17}* & 63.60{\scriptsize \color{grayv} $\pm$0.64}* & 73.42{\scriptsize \color{grayv} $\pm$1.18}* & 68.76{\scriptsize \color{grayv} $\pm$1.12}* & 70.49{\scriptsize \color{grayv} $\pm$1.06}* & \textbf{+1.41} \\

    BMR \cite{ying2023bootstrapping} & 67.12{\scriptsize \color{grayv} $\pm$0.74} & 66.64{\scriptsize \color{grayv} $\pm$1.28} & 59.09{\scriptsize \color{grayv} $\pm$0.61} & 72.62{\scriptsize \color{grayv} $\pm$1.28} & 64.43{\scriptsize \color{grayv} $\pm$1.28} & 75.10{\scriptsize \color{grayv} $\pm$1.13} & 62.56{\scriptsize \color{grayv} $\pm$0.91} & 68.65{\scriptsize \color{grayv} $\pm$1.17} & - \\
    \rowcolor{lightgrayv} \quad + \baby & 67.84{\scriptsize \color{grayv} $\pm$0.83}* & 67.68{\scriptsize \color{grayv} $\pm$0.82}* & 60.01{\scriptsize \color{grayv} $\pm$0.88}* & 73.31{\scriptsize \color{grayv} $\pm$1.28}* & 65.65{\scriptsize \color{grayv} $\pm$0.92}* & 76.27{\scriptsize \color{grayv} $\pm$1.03}* & 64.32{\scriptsize \color{grayv} $\pm$0.98}* & 69.71{\scriptsize \color{grayv} $\pm$0.91}* & \textbf{+1.08} \\
    \bottomrule
  \end{tabular} }
\end{table*}

\noindent
\textbf{Baselines.}
Across three datasets \textit{GossipCop} \cite{shu2020fakenewsnet}, \textit{Weibo} \cite{jin2017multimodal}, \textit{Twitter} \cite{boididou2018detection} ($K=1$), we compare five MMD baselines and their improved versions using \baby in our experiments. A brief overview of these baseline models is provided below:
\begin{itemize}
    \item \textbf{Base model}. The baseline framework extracts textual and visual features, projects them into a shared representation space via feed-forward neural network layers for semantic alignment, and subsequently performs feature fusion. The fused features are fed into multi-layer perceptron layers to produce the final veracity classification.
    \item \textbf{SAFE} \cite{zhou2020safe}. This method introduces a multimodal fusion module that considers semantic similarity across modalities to enhance feature integration for MMD.
    \item \textbf{MCAN} \cite{wu2021multimodal}. It utilizes a co-attention mechanism that integrates multimodal features by accounting for inter-modality relationships
    \item \textbf{CAFE} \cite{chen2022cross}. This approach adopts a variational weighting strategy to dynamically control multimodal feature fusion, and applies contrastive learning to enhance cross-modal feature alignment.
    \item \textbf{BMR} \cite{ying2023bootstrapping}. It constructs an advanced network with an improved mixture-of-experts mechanism for both feature extraction and fusion of multimodal content.
\end{itemize}

For all baselines, we re-produced them by employing the BERT model and ResNet34 as the feature extractors. Additionally, across the \textit{FakeSV} dataset \cite{qi2023fakesv} ($K \geq 2$), we compare the following four baselines:
\begin{itemize}
    \item \textbf{VGG} \cite{simonyan2015very} extracts visual features of video frames with the pre-trained VGG-19 network.
    \item \textbf{C3D} \cite{tran2015learning} is a pre-trained video analysis model, which captures temporal and motion information from sequences of video frames.
    \item \textbf{FANVM} \cite{choi2021using} learns the topic inconsistency between the text content and the comments with an adversarial network, and integrates them with the frame features.
    \item \textbf{SV-FEND} \cite{qi2023fakesv} extracts text, audio, frame, comment, and user features with different encoders, and fuses them with a cross-modal transformer model.
\end{itemize}

\noindent
\textbf{Implementation Details.}
During data preprocessing, raw images are resized and randomly cropped to 224 $\times$ 224 pixels, and text content is limited to 128 tokens. We then use pre-trained ResNet34\footnote{\url{https://download.pytorch.org/models/resnet34-333f7ec4.pth}.} and BERT\footnote{\url{https://huggingface.co/bert-base-uncased}.} models to extract visual and textual features, keeping the first 9 layers of BERT’s Transformer frozen. For audio features from video content, we employ a pre-trained VGGish model \cite{hershey2017cnn}.
The manipulation teacher model is based on a shallow ResNet18 architecture, which is pre-trained on the IMD benchmark dataset CASIAv2\footnote{\url{https://github.com/SunnyHaze/CASIA2.0-Corrected-Groundtruth}.} \cite{dong2013casia,pham2019hybrid}, comprising 12,614 images, with 7,491 authentic and 5,123 tampered examples.
In the training phase, we fine-tune the BERT model using the Adam optimizer with a learning rate of $3 \times 10^{-5}$, while other modules are optimized separately with Adam at a learning rate of $10^{-3}$, using a batch size of 32 throughout. The hyperparameters $\alpha$, $\beta$, $\delta$, and $K$ are set to 0.1, 0.1, 0.1, and 10, respectively. To prevent overfitting, training stops if the Macro F1 score does not improve for 10 epochs. Each experiment is run 5 times with different random seeds ${1, 2, 3, 4, 5}$, and we report the average scores from these trials in the final results.

\subsection{Main Results}

We evaluate our proposed model, \baby, against nine baseline approaches on four benchmark datasets, with the results summarized in Tables~\ref{result} and \ref{resultfakesv}. These tables report the Avg. $\Delta$ scores, which quantify the average performance gains achieved by \baby over each baseline across all evaluation metrics. Overall, our model exhibits substantial improvements against all competitors. For instance, on the Weibo dataset, \baby surpasses BMR by approximately 1.79 points, while on Twitter it exceeds the Base model by 1.84 points.
A more fine-grained analysis of \baby's performance across various metrics highlights its consistent superiority over the baselines. For example, on the \textit{Gossipcop} dataset, it surpasses the BMR model by around 2.57 in the fake class recall score. These outcomes underscore the efficacy of our approach and emphasize the significance of manipulation and intention features in misinformation detection.
Furthermore, we note that the magnitude of improvements by \baby across the four MMD datasets roughly follows the order \textit{FakeSV} $>$ \textit{Twitter} $>$ \textit{Weibo} $>$ \textit{GossipCop}. This pattern suggests that \baby yields larger benefits on smaller datasets, where the scarcity of semantic information can be compensated by leveraging manipulation and intention cues. Furthermore, the relatively high prevalence of manipulated images in the Twitter dataset amplifies the contribution of manipulation-aware features, indirectly validating the reliability and relevance of the extracted manipulation

\begin{table*}
\centering
\renewcommand\arraystretch{1.1}
  \caption{Experimental results of \baby across the prevalent MMD dataset \textit{FakeSV}. The results marked by * are statistically significant compared to its baseline models, satisfying p-value $<$ 0.05.}
  \label{resultfakesv}
  \setlength{\tabcolsep}{5pt}{
  \begin{tabular}{m{2.2cm}m{1.50cm}<{\centering}m{1.50cm}<{\centering}m{1.50cm}<{\centering}m{1.50cm}<{\centering}m{1.50cm}<{\centering}m{1.20cm}<{\centering}}
    \toprule
    \quad \quad Method & Accuracy & Macro F1 & Precision & Recall & AUC & Avg. $\Delta$\\
    \hline
    VGG \cite{simonyan2015very} & 69.45{\scriptsize \color{grayv} $\pm$2.12} & 64.52{\scriptsize \color{grayv} $\pm$1.96} & 68.08{\scriptsize \color{grayv} $\pm$2.87} & 64.19{\scriptsize \color{grayv} $\pm$1.84} & 64.19{\scriptsize \color{grayv} $\pm$1.84} & - \\
    \rowcolor{lightgrayv} \quad + \baby & 70.85{\scriptsize \color{grayv} $\pm$1.31}* & 66.12{\scriptsize \color{grayv} $\pm$2.35}* & 70.00{\scriptsize \color{grayv} $\pm$0.78}* & 65.82{\scriptsize \color{grayv} $\pm$1.91}* & 65.82{\scriptsize \color{grayv} $\pm$1.91}* & \textbf{+1.63} \\

    C3D \cite{tran2015learning} & 67.60{\scriptsize \color{grayv} $\pm$1.62} & 67.43{\scriptsize \color{grayv} $\pm$1.59} & 67.97{\scriptsize \color{grayv} $\pm$1.67} & 67.60{\scriptsize \color{grayv} $\pm$1.58} & 67.60{\scriptsize \color{grayv} $\pm$1.58} & - \\
    \rowcolor{lightgrayv} \quad + \baby & 68.80{\scriptsize \color{grayv} $\pm$1.77}* & 68.70{\scriptsize \color{grayv} $\pm$1.79}* & 68.98{\scriptsize \color{grayv} $\pm$1.77}* & 68.77{\scriptsize \color{grayv} $\pm$1.78}* & 68.77{\scriptsize \color{grayv} $\pm$1.78}* & \textbf{+1.16} \\

    FANVM \cite{choi2021using} & 77.14{\scriptsize \color{grayv} $\pm$3.07} & 77.11{\scriptsize \color{grayv} $\pm$3.05} & 77.26{\scriptsize \color{grayv} $\pm$3.14} & 77.13{\scriptsize \color{grayv} $\pm$3.04} & 77.13{\scriptsize \color{grayv} $\pm$3.04} & - \\
    \rowcolor{lightgrayv} \quad + \baby & 78.39{\scriptsize \color{grayv} $\pm$2.44}* & 78.38{\scriptsize \color{grayv} $\pm$2.42}* & 78.48{\scriptsize \color{grayv} $\pm$2.54}* & 78.39{\scriptsize \color{grayv} $\pm$2.43}* & 78.39{\scriptsize \color{grayv} $\pm$2.43}* & \textbf{+1.25} \\

    SV-FEND \cite{qi2023fakesv} & 79.03{\scriptsize \color{grayv} $\pm$2.09} & 78.99{\scriptsize \color{grayv} $\pm$2.23} & 79.94{\scriptsize \color{grayv} $\pm$1.30} & 79.17{\scriptsize \color{grayv} $\pm$2.00} & 79.17{\scriptsize \color{grayv} $\pm$2.00} & - \\
    \rowcolor{lightgrayv} \quad + \baby & 80.67{\scriptsize \color{grayv} $\pm$1.84}* & 80.63{\scriptsize \color{grayv} $\pm$1.80}* & 80.91{\scriptsize \color{grayv} $\pm$2.14}* & 80.66{\scriptsize \color{grayv} $\pm$1.82}* & 80.66{\scriptsize \color{grayv} $\pm$1.82}* & \textbf{+1.44} \\
    \bottomrule
  \end{tabular} }
\end{table*}

\subsection{Ablative Study}

To assess how varying objective functions and features influence our model, \baby, an ablation study was conducted on three distinct datasets: the English-language \textit{GossipCop}, the Chinese-language \textit{Weibo}, and the video-oriented \textit{FakeSV} dataset. Tables~\ref{ablative.loss} and \ref{ablative.feature} display the results of these experiments. Specific descriptions of the different ablated versions are presented as follows:
\begin{itemize}
    \item \textbf{w/o} $\mathcal{L}_{PRE}$. This variant trains the teacher model solely using the PU objective $\mathcal{L}_{PU}$, omitting the initial pre-training phase involving the external IMD dataset $\mathcal{D}_\mu$;
    \item \textbf{w/o} $\mathcal{L}_{PU}$. In this case, the PU loss is excluded from adapting the teacher model to the MMD dataset  $\mathcal{D}$;
    \item \textbf{w/o} $\mathcal{L}_{KD}, \mathcal{L}_{IR}$. In this variant, no objective function is applied to control the manipulation and intention features. Instead, the cross-entropy loss $\mathcal{L}_{CE}$ alone is used for veracity prediction. Since $\mathcal{L}_{IR}$ is highly related to $\mathcal{L}_{KD}$, both are removed simultaneously.
    \item \textbf{w/o} $\mathbf{e}^m$, \textbf{w/o} $\mathbf{e}^e$, and \textbf{w/o} $\mathbf{e}^m, \mathbf{e}^e$. These variations exclude the features $\mathbf{e}^m$ and $\mathbf{e}^e$ and their corresponding training losses. Removing both features leads to a model that aligns with the baseline, omitting the enhancements introduced by \baby in this study.
\end{itemize}

\begin{table*}[t]
\centering
\renewcommand\arraystretch{1.2}
  \caption{Ablative study on objective functions. The bold and underlined scores indicate the highest and lowest results in the ablative versions, respectively.}
  \label{ablative.loss}
  \setlength{\tabcolsep}{5pt}{
  \begin{tabular}{m{2.40cm}m{0.62cm}<{\centering}m{0.62cm}<{\centering}m{0.62cm}<{\centering}m{0.68cm}<{\centering}m{0.68cm}<{\centering}m{0.62cm}<{\centering}m{0.62cm}<{\centering}m{0.62cm}<{\centering}m{0.68cm}<{\centering}m{0.68cm}<{\centering}m{0.62cm}<{\centering}m{0.62cm}<{\centering}m{0.62cm}<{\centering}m{0.62cm}<{\centering}m{0.62cm}<{\centering}}
    \toprule
    \multirow{2}{*}{\quad Method} & \multicolumn{5}{c}{\textbf{Dataset}: \textit{GossipCop}} & \multicolumn{5}{c}{\textbf{Dataset}: \textit{Weibo}} & \multicolumn{5}{c}{\textbf{Dataset}: \textit{FakeSV}} \\
    \cmidrule (r){2-6} \cmidrule (r){7-11} \cmidrule (r){12-16}
    & Acc. & F1 & AUC & F1$_\text{real}$ & F1$_\text{fake}$ & Acc. & F1 & AUC & F1$_\text{real}$ & F1$_\text{fake}$ & Acc. & F1 & P. & R. & AUC \\
    \hline
    \rowcolor{lightgrayv} SOTA + \baby & 87.95 & 79.99 & 87.46 & 93.14 & 66.80 & 91.74 & 91.68 & 97.06 & 91.56 & 91.81 & 80.67 & 80.63 & 80.91 & 80.66 & 80.66 \\
    \ w/o $\mathcal{L}_{PRE}$ & 87.24 & 79.18 & 87.21 & 92.14 & 66.23 & 90.17 & 90.12 & 96.93 & 89.91 & 90.83 & 79.45 & 79.41 & 79.70 & 79.47 & 79.47 \\
    \ w/o $\mathcal{L}_{PU}$ & \textbf{87.63} & \textbf{79.66} & \textbf{87.38} & \textbf{93.05} & \textbf{66.52} & \textbf{91.40} & \textbf{91.39} & \textbf{96.96} & \textbf{91.19} & \textbf{91.60} & \textbf{80.35} & \textbf{80.32} & \textbf{80.52} & \textbf{80.36} & \textbf{80.36} \\
    \ w/o $\mathcal{L}_{KD}, \mathcal{L}_{IR}$ & \underline{86.93} & \underline{78.65} & \underline{86.24} & \underline{91.94} & \underline{65.36} & \underline{90.10} & \underline{90.10} & \underline{95.97} & \underline{89.88} & \underline{90.31} & \underline{78.98} & \underline{78.94} & \underline{79.21} & \underline{79.00} & \underline{79.00} \\
    \bottomrule
  \end{tabular} }
\end{table*}

\begin{table*}[t]
\centering
\renewcommand\arraystretch{1.2}
  \caption{Ablative study on manipulation and intention features.}
  \label{ablative.feature}
  \setlength{\tabcolsep}{5pt}{
  \begin{tabular}{m{2.40cm}m{0.62cm}<{\centering}m{0.62cm}<{\centering}m{0.62cm}<{\centering}m{0.68cm}<{\centering}m{0.68cm}<{\centering}m{0.62cm}<{\centering}m{0.62cm}<{\centering}m{0.62cm}<{\centering}m{0.68cm}<{\centering}m{0.68cm}<{\centering}m{0.62cm}<{\centering}m{0.62cm}<{\centering}m{0.62cm}<{\centering}m{0.62cm}<{\centering}m{0.62cm}<{\centering}}
    \toprule
    \multirow{2}{*}{\quad Method} & \multicolumn{5}{c}{\textbf{Dataset}: \textit{GossipCop}} & \multicolumn{5}{c}{\textbf{Dataset}: \textit{Weibo}} & \multicolumn{5}{c}{\textbf{Dataset}: \textit{FakeSV}} \\
    \cmidrule (r){2-6} \cmidrule (r){7-11} \cmidrule (r){12-16}
    & Acc. & F1 & AUC & F1$_\text{real}$ & F1$_\text{fake}$ & Acc. & F1 & AUC & F1$_\text{real}$ & F1$_\text{fake}$ & Acc. & F1 & P. & R. & AUC \\
    \hline
    \rowcolor{lightgrayv} SOTA + \baby & 87.95 & 79.99 & 87.46 & 93.14 & 66.80 & 91.74 & 91.68 & 97.06 & 91.56 & 91.81 & 80.67 & 80.63 & 80.91 & 80.66 & 80.66 \\
    \ w/o $\mathbf{e}^m$ & 87.53 & 79.13 & 86.47 & 92.37 & 65.89 & 90.92 & 90.91 & 96.52 & 90.57 & 91.08 & 79.77 & 79.72 & 79.93 & 79.79 & 79.79 \\
    \ w/o $\mathbf{e}^e$ & \textbf{87.69} & \textbf{79.56} & \textbf{86.72} & \textbf{92.39} & \textbf{66.25} & \textbf{91.13} & \textbf{91.11} & \textbf{96.78} & \textbf{90.78} & \textbf{91.45} & \textbf{80.29} & \textbf{80.23} & \textbf{80.40} & \textbf{80.26} & \textbf{80.26} \\
    \ w/o $\mathbf{e}^m, \mathbf{e}^e$ & \underline{87.26} & \underline{79.03} & \underline{86.27} & \underline{92.14} & \underline{65.51} & \underline{90.17} & \underline{90.15} & \underline{96.45} & \underline{89.81} & \underline{90.50} & \underline{79.03} & \underline{78.99} & \underline{79.94} & \underline{79.17} & \underline{79.17} \\
    \bottomrule
  \end{tabular} }
\end{table*}

Overall, the removal of each component weakens \baby's predictive performance, highlighting the importance of each element. Notably, the predictive performance of the three ablation objectives ranks in the order of w/o $\mathcal{L}_{PU}$ $>$ w/o $\mathcal{L}_{PRE}$ $>$ w/o $\mathcal{L}_{KD}, \mathcal{L}_{IR}$. Excluding $\mathcal{L}_{PRE}$ and $\mathcal{L}_{KD}, \mathcal{L}_{IR}$ particularly diminishes the model’s performance, sometimes even falling below that of the baseline.
When $\mathcal{L}_{PRE}$ is removed, the teacher’s predictive power declines, resulting in less distinct features produced by the manipulation encoder, which adversely affects veracity predictions. In contrast, the absence of $\mathcal{L}_{KD}$ and $\mathcal{L}_{IR}$ leads to unregulated manipulation and intention features, increasing the computational burden of the baseline model and introducing redundant, non-discriminative features that weaken the discriminative strength of the final veracity feature $\mathbf{e}$.
Comparing the performance impacts of three variants by the feature removal, we find the ranking as w/o $\mathbf{e}^e$ $>$ w/o $\mathbf{e}^m$ $>$ w/o $\mathbf{e}^m, \mathbf{e}^e$, underscoring the value of both features in enhancing the final multimodal feature’s discriminative ability, with $\mathbf{e}^m$ making a stronger contribution.
Specifically, after removing the manipulation feature $\mathbf{e}^m$, the model's performance decreases more significantly. This observation first demonstrates that our method can extract accurate and discriminative manipulation features. It indirectly validates that our proposed manipulation teacher model can enhance its generalization across various manipulation types by simultaneously performing incremental learning on external IMD data and PU learning on synthetic MMD data. Additionally, it effectively transfers the pre-trained model from IMD data to MMD data, thereby mitigating the data shift problem.

\begin{figure*}[t]
  \centering
  \includegraphics[scale=0.19]{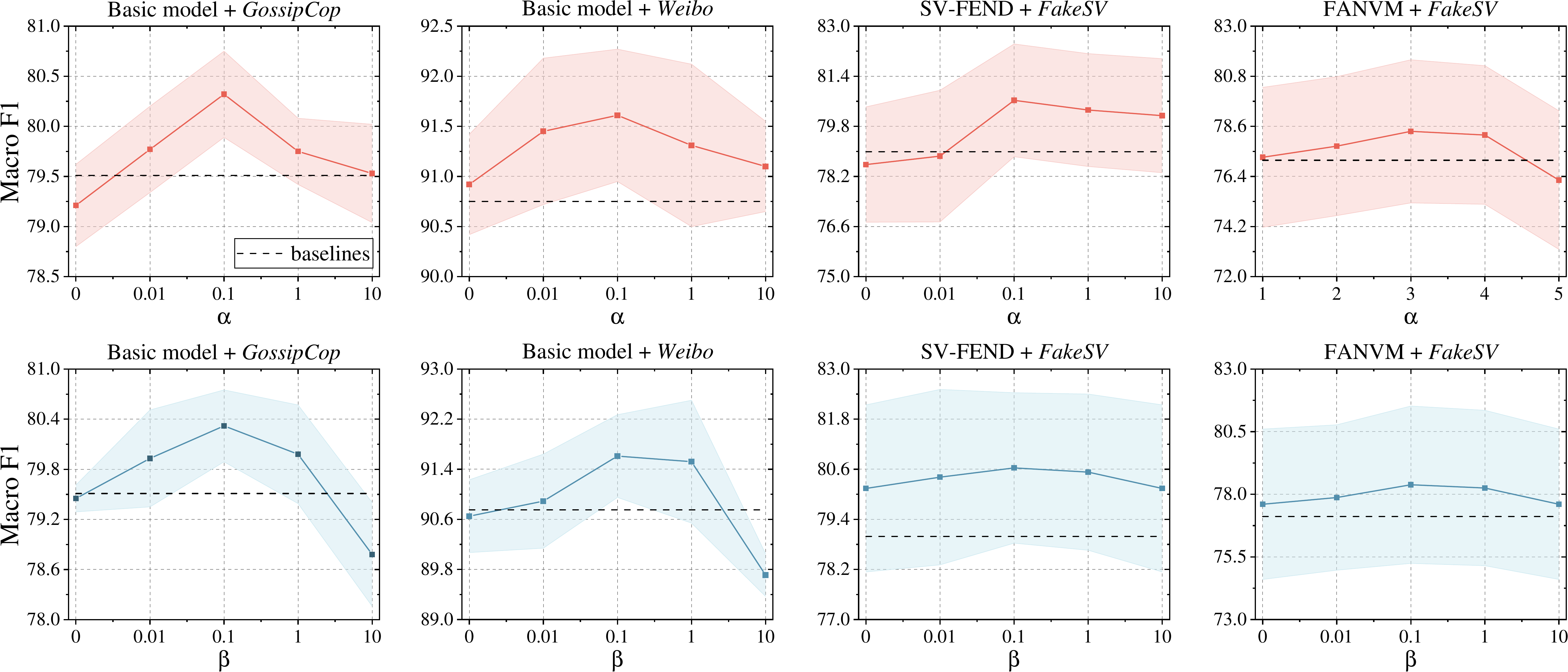}
  \caption{Sensitivity analysis of the parameters $\alpha$ and $\beta$.}
  \label{sensitivity}
\end{figure*}

\subsection{Sensitivity Analysis}

In the \baby method, the hyper-parameters $\alpha$ and $\beta$ play crucial roles by determining the relative importance of $\mathcal{L}_{KD}$ and $\mathcal{L}_{IR}$ during training, thus helping to maintain a balance across the various loss functions. To evaluate the model's sensitivity to these hyper-parameters, we conduct sensitivity analyses, and the corresponding experimental results are shown in Fig.~\ref{sensitivity}.
We conduct experiments on four versions: Basic model + \textit{GossipCop}, Basic model + \textit{Weibo}, SV-FEND + \textit{FakeSV}, and FANVM + \textit{FakeSV}, with the Macro F1 metric reported in Fig.~\ref{sensitivity}.
$\alpha$ and $\beta$ are drawn from the set $\{0, 0.01, 0.1, 1, 10\}$, where $\alpha$ or $\beta = 0$ indicates the corresponding objective function is not engaged in training.
The results indicate that the model is highly sensitive to both hyper-parameters, performing optimally when $\alpha = 0.1$ and $\beta = 0.1$. Deviating from these values in either direction leads to diminished performance. Consequently, we consistently use $\alpha = 0.1$ and $\beta = 0.1$ for all experiments in this study.
When $\alpha$ and $\beta$ are too small, the manipulation and intention features are undertrained, resulting in inadequate discriminative features that impair the model’s predictive accuracy, potentially falling short of the baseline model. On the other hand, if they are too large, the model’s optimization prioritizes their respective loss functions, diminishing the weight of the veracity prediction objective $\mathcal{L}_{CE}$ and negatively impacting the veracity prediction results.

\begin{figure}[t]
  \centering
  \includegraphics[scale=0.43]{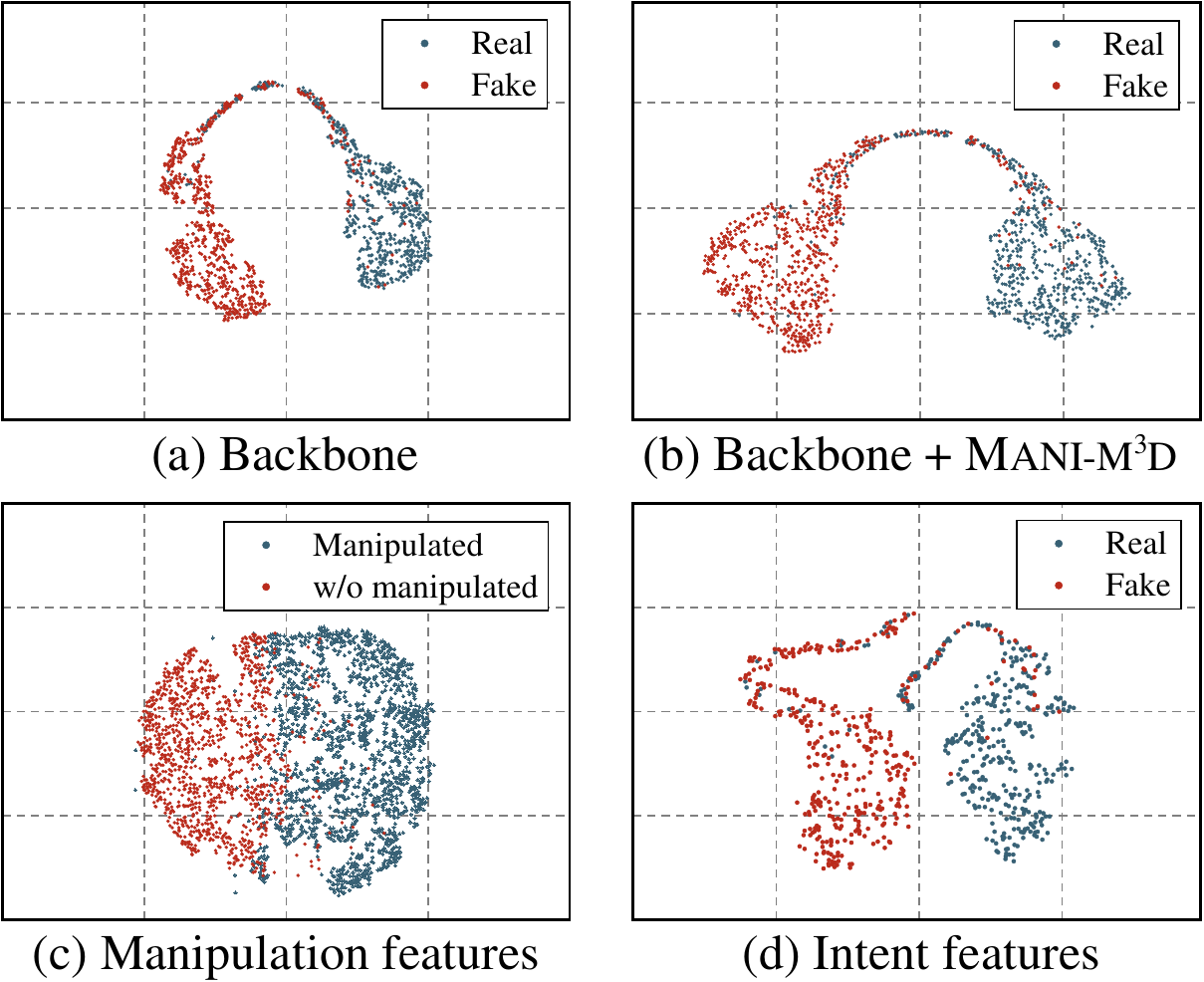}
  \caption{Visualization analysis of features $\mathbf{z}$, $\mathbf{e}^m$ and $\mathbf{e}^e$ with the T-SNE method.}
  \label{visualization}
\end{figure}

\subsection{Visualization Analysis}

To investigate the discriminative capability of the extracted manipulation and intention features, we provide a visualization analysis of these features in Fig.~\ref{visualization}. We select the \textit{Weibo} dataset for this visualization analysis and employ the T-SNE technique \cite{van2008visualizing} to reduce the multimodal feature $\mathbf{z}$, the manipulation feature $\mathbf{e}^m$, and the intention feature $\mathbf{e}^e$ to a 2D space, visualizing the corresponding points in Fig.~\ref{visualization}.  Fig.~\ref{visualization}(a) depicts the visualization of the multimodal features of the basic model, while Fig.~\ref{visualization}(b) presents the visualization of the multimodal features when enhanced by our proposed \baby. A comparison of the two figures reveals that our method effectively separates the clusters of \textit{real} and \textit{fake} classes, thereby enhancing the discriminative nature of the multimodal features.
Fig.~\ref{visualization}(c) shows the visualization of the manipulation feature $\mathbf{e}^m$, where we utilize the teacher model’s results to differentiate between \textit{manipulated} and \textit{unmanipulated} images. The visualization demonstrates the strong discriminative power of the manipulation feature in identifying manipulated images, showcasing the effectiveness of knowledge distillation in transferring the discriminative ability of the manipulation teacher to the manipulation encoder. Meanwhile, this result on the discriminative capability of manipulation features directly demonstrates that our trained manipulation teacher model can enhance its generalization to different and even unknown visual content manipulation types through online learning on IMD data. It also shows that the pre-trained teacher model on IMD data can be effectively transferred to MMD data by using synthetic copy-moved data.
Fig.~\ref{visualization}(d) focuses on the intention features $\mathbf{e}^e$ of samples classified as “\textit{manipulated}” in the manipulation classification task. We observe a clear separation between the intention features of \textit{real} and \textit{fake} samples. However, some \textit{fake} samples are interspersed within the \textit{real} cluster, indicating that \textit{fake} samples may also exhibit \textit{harmless} intention.

\begin{figure}[t]
  \centering
  \includegraphics[scale=0.48]{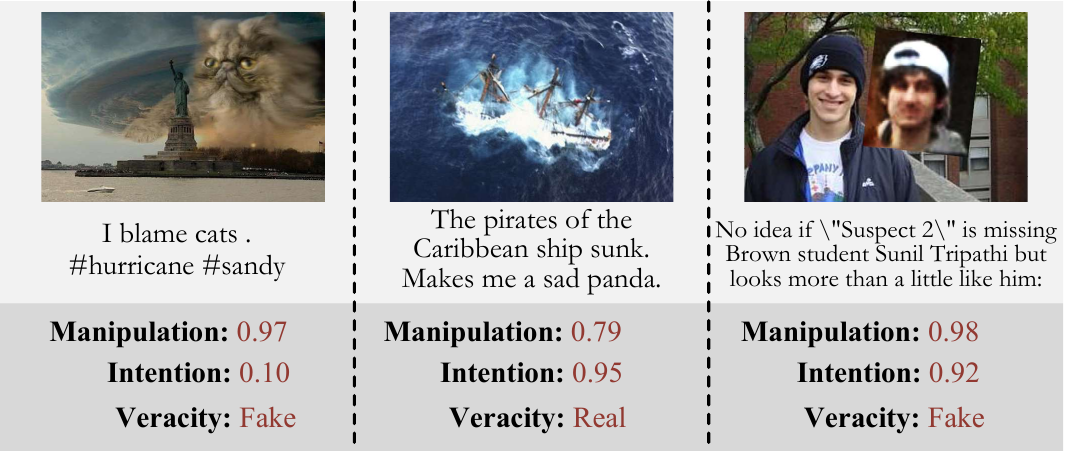}
  \caption{We illustrate three representative examples for the case study.}
  \label{case}
\end{figure}

\subsection{Case Study}
We present three illustrative cases in Fig.~\ref{case} to showcase the efficacy of our classifiers in handling the three different tasks. In the first case, we examine an image that exhibits manipulation with harmful intentions. Our model successfully detects both the manipulation and the harmful intention, correctly labeling it as fake. 
The second case shows an image with manipulated colors. While our model correctly identifies the manipulation and predicts its veracity label with precision, it assigns a low-confidence score to this specific manipulation type, indicating the potential for enhancement in recognizing certain manipulation methods. The third case involves an image that has undergone harmless manipulation, merely being cropped without any harmful intention. Our model demonstrates accurate predictions across all three tasks. Overall, our model exhibits impressive performance in the three tasks, which indicates the effectiveness of our manipulation detector distilled from the manipulation teacher and the intention detector by the heuristic PU learning. However, it remains susceptible to subtle manipulations, highlighting the need for further refinement.

\subsection{Computation Budget}

Due to the introduction of an external IMD dataset for training the manipulation teacher model and the proposal of two manipulation and intention detectors to generate their features, this process undoubtedly imposes a greater computational burden and reduces computational efficiency. To quantify the computational burden introduced by our method, we conduct a time-cost experiment. We test the time consumption for a single training session of nine baseline models across four MMD datasets, both with and without our proposed method. Since we employ an early stopping strategy during training, which can significantly vary the training time, each experiment is repeated five times with different seeds \{1,2,3,4,5\}, and the average training time is reported in Table~\ref{budget}.

The experimental results show that, after applying \baby, the time consumption increased by approximately 1.27 times compared to the baseline methods. This additional time is primarily due to the training of the manipulation teacher model using external IMD data. However, our ablation studies indicate that the manipulation features obtained from this training significantly enhance the model's performance. Therefore, the additional time overhead is justifiable and acceptable.

\begin{table}[t]
\centering
\renewcommand\arraystretch{1.1}
  \caption{Training budget of our proposed \baby.}
  \label{budget}
  \setlength{\tabcolsep}{5pt}{
  \begin{tabular}{m{1.7cm}<{\centering}m{0.85cm}<{\centering}m{0.85cm}<{\centering}m{0.85cm}<{\centering}m{1.2cm}<{\centering}m{0.85cm}<{\centering}}
    \toprule
    Methods & Base & SAFE & MCAN & CAFE & BMR \\

    \hline
    \textit{GossipCop} & 20.2 & 23.0 & 28.8 & 18.8 & 23.8 \\
    + \baby & 27.4 & 27.8 & 34.6 & 23.6 & 27.8 \\
    \rowcolor{lightgrayv} $\Delta$ & \textbf{1.35}$\times$ & \textbf{1.20}$\times$ & \textbf{1.20}$\times$ & \textbf{1.25}$\times$ & \textbf{1.16}$\times$ \\
    
    \hline
    \textit{Weibo} & 8.0 & 10.0 & 15.8 & 10.0 & 10.6 \\
    + \baby & 12.0 & 14.2 & 18.8 & 13.8 & 14.0 \\
    \rowcolor{lightgrayv} $\Delta$ & \textbf{1.50}$\times$ & \textbf{1.42}$\times$ & \textbf{1.18}$\times$ & \textbf{1.38}$\times$ & \textbf{1.32}$\times$ \\
    
    \hline  
    \textit{Twitter} & 15.8 & 14.8 & 29.2 & 16.0 & 20.4 \\
    + \baby & 20.4 & 19.0 & 34.4 & 19.8 & 25.8 \\
    \rowcolor{lightgrayv} $\Delta$ & \textbf{1.29}$\times$ & \textbf{1.28}$\times$ & \textbf{1.17}$\times$ & \textbf{1.23}$\times$ & \textbf{1.26}$\times$ \\
    
    \hline
    \specialrule{0em}{0.5pt}{0.5pt}
    \hline
    Methods & VGG & C3D & FANVM & SV-FEND &  \\
    \hline
    \textit{FakeSV} & 10.2 & 13.8 & 27.4 & 73.0 &  \\
    + \baby & 14.2 & 18.4 & 32.8 & 78.4 &  \\
    \rowcolor{lightgrayv} $\Delta$ & \textbf{1.39}$\times$ & \textbf{1.33}$\times$ & \textbf{1.19}$\times$ & \textbf{1.07}$\times$ &  \\
    \bottomrule
  \end{tabular} }
\end{table}

\section{Conclusion and Future Works}

In this work, we focus on detecting multimodal misinformation by identifying traces of manipulation in visual content within articles, along with understanding the underlying intentions behind such manipulations. To achieve this, we present a novel MMD model called \baby, which extracts manipulation and intention features, then integrates them into the overall multimodal features. To enhance the discriminative capability of these features regarding whether an image has been harmfully manipulated, we propose two classifiers that predict the respective labels. To address the issue of unknown manipulation and intention labels, we introduce two weakly supervised signals. These signals are learned by training a manipulation teacher on additional IMD datasets, along with two PU learning objectives that adapt and guide the classifier. Our experimental results demonstrate that \baby significantly outperforms baseline models.

\vspace{2pt} \noindent
\textbf{Future works.}
First, a more powerful and meticulously designed manipulation detection teacher can improve the MMD performance, and in our future work, we aim to gradually enhance the manipulation detection performance by incorporating more diverse IMD data and refining the model architecture. However, introducing more data and more complex models invariably leads to increased time consumption. Balancing the additional training costs with model performance will be a key focus of future work.
Then, the continuous updates on social media platforms often expose models to previously unseen events or emerging domains. To address this emergent issue, our method builds upon existing advanced MMD techniques by introducing additional image manipulation features and manipulation intention features as supplementary cues, and offers greater flexibility to incorporate advanced data adaptation techniques when emergent events significantly deviate from existing data. However, dealing with unseen events is always a challenging problem, as even humans struggle to assess the veracity of an event accurately. Addressing this issue will be a key focus of our future work.

\bibliographystyle{IEEEtran}
\bibliography{reference}

\vspace{-10pt}
\begin{IEEEbiography}[{\includegraphics[width=1in,height=1.25in,clip,keepaspectratio]{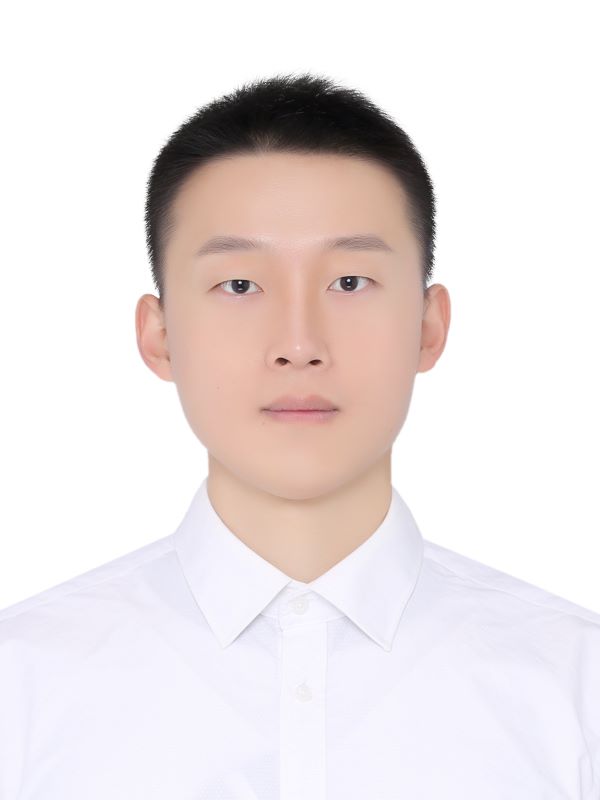}}]{Bing Wang}
received the MS degree in computer science from Jilin University, Changchun, China, in 2023 and the BS degree in industrial engineering from Jilin University in 2020. He is currently pursuing the PhD degree in computer science from Jilin University. He has published more than 20 papers in international journals and conferences, including SIGIR, ACM Multimedia, IJCAI, ICML, NeurIPS, AAAI, CIKM, \etc His research interest is large language models and misinformation detection.
\end{IEEEbiography}

\vspace{-10pt}
\begin{IEEEbiography}[{\includegraphics[width=1in,height=1.25in,clip,keepaspectratio]{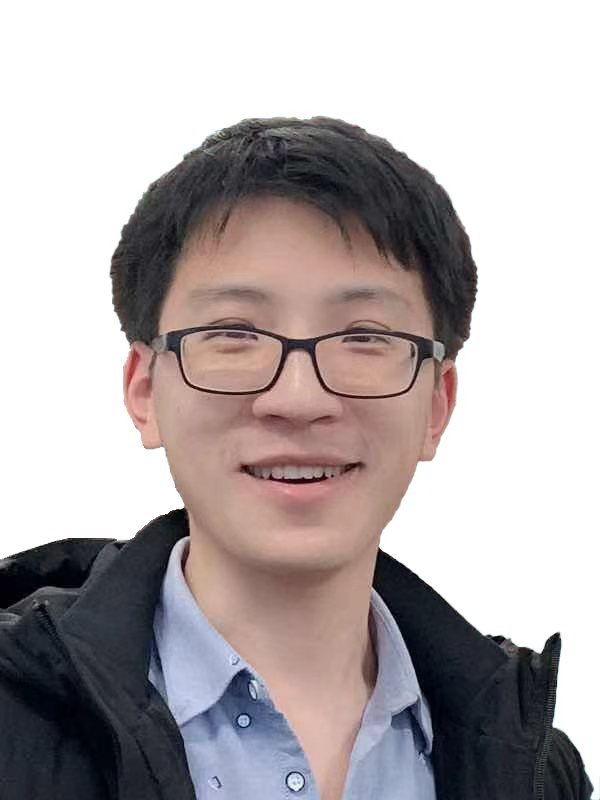}}]{Ximing Li}
received the Ph.D. degree from Jilin University, Changchun, China, in 2015. 
He is currently a Professor with the College of Computer Science and Technology, Jilin University, Changchun, China. His research interests include data mining, machine learning and natural language processing. He has published more than 100 papers at the competitive venues, including NeurIPS, ICML, ICLR, ACL, IJCAI, AAAI, WWW \etc
\end{IEEEbiography}

\vspace{-10pt}
\begin{IEEEbiography}[{\includegraphics[width=1in,height=1.25in,clip,keepaspectratio]{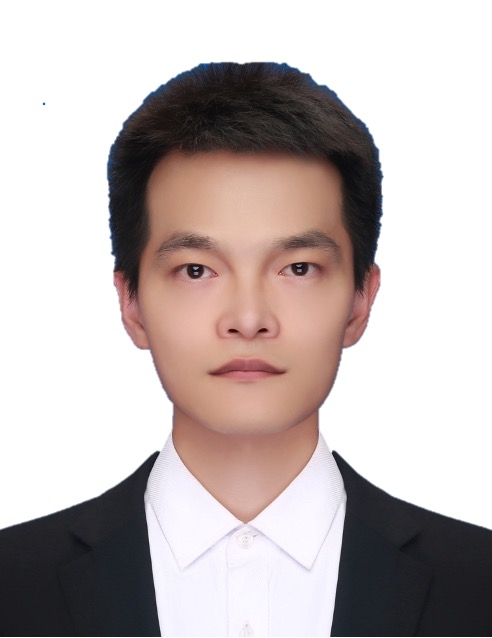}}]{Changchun Li}
received the Ph.D. degree from Jilin University, Changchun, China, in 2022. 
He is currently an Associate Professor with the College of Computer Science and Technology, Jilin University, Changchun, China. His research interests include data mining and machine learning, especially for weakly supervised learning and semi-supervised learning. He has published many papers at the competitive venues, including NeurIPS, ICML, ICLR, ACL, AAAI, WWW, CIKM, ICDM \etc
\end{IEEEbiography}

\vspace{-10pt}
\begin{IEEEbiography}[{\includegraphics[width=1in,height=1.25in,clip,keepaspectratio]{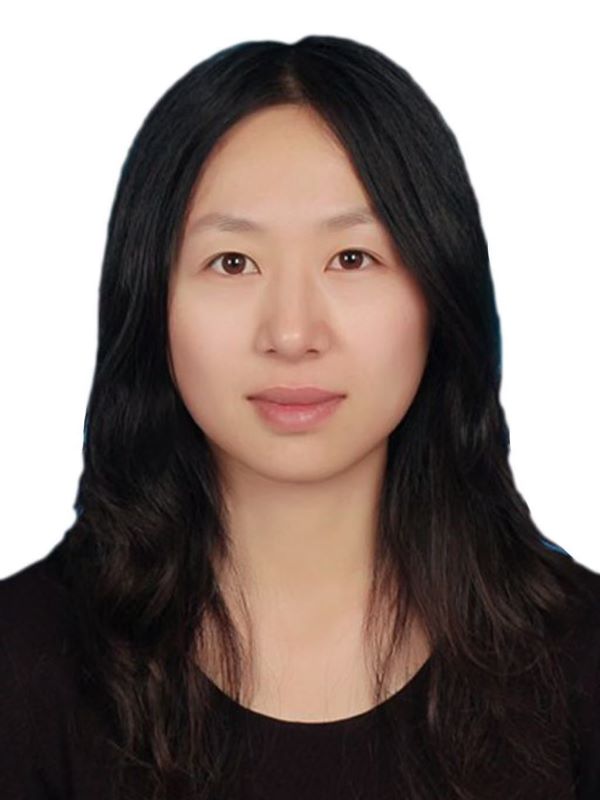}}]{Jinjin Chi}
received the Ph.D. degree from Jilin University, Changchun, China, in 2019. 
She is currently an Associate Professor with the College of Computer Science and Technology, Jilin University. His research interests include optimal transport and representation learning. She has published papers at the competitive venues, including IJCAI, AAAI \etc
\end{IEEEbiography}

\vspace{-10pt}
\begin{IEEEbiography}[{\includegraphics[width=1in,height=1.25in,clip,keepaspectratio]{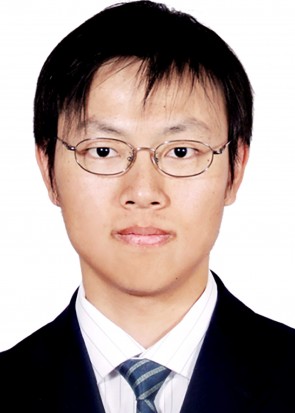}}]{Renchu Guan}
received the Ph.D. degree from the College of Computer Science and Technology, Jilin University, Changchun, China, in 2010. He was a Visiting Scholar with the University of Trento, Trento, Italy, from 2011 to 2012. He is currently a Professor with the College of Computer Science and Technology, Jilin University. He has authored or coauthored over 40 scientific papers in refereed journals and proceedings. His research was featured in the Nature Communications, IEEE Transactions on Knowledge and Data Engineering, IEEE Transactions on Geoscience and Remote Sensing, \etc His research interests include machine learning and knowledge engineering.
\end{IEEEbiography}

\vspace{-10pt}
\begin{IEEEbiography}[{\includegraphics[width=1in,height=1.25in,clip,keepaspectratio]{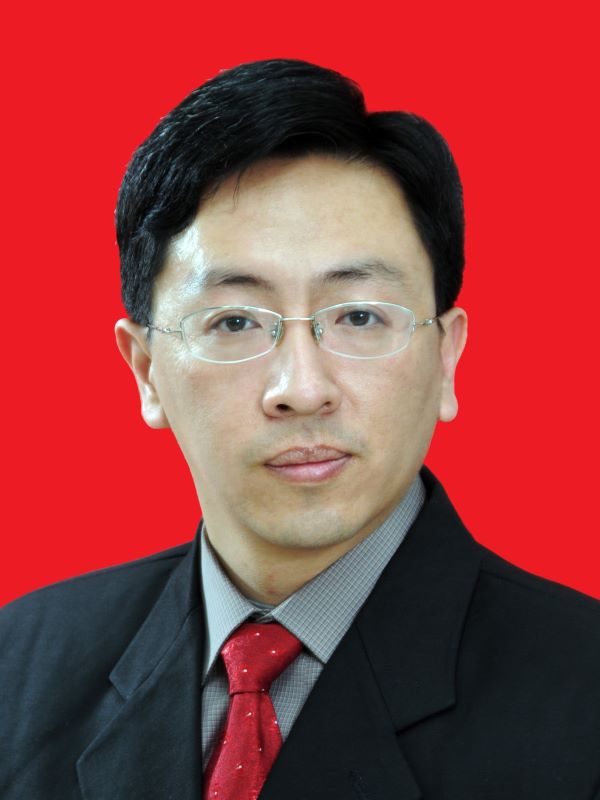}}]{Shengsheng Wang}
received the B.S., M.S., and Ph.D. degrees in computer science from Jilin University, Changchun, China, in 1997, 2000, and 2003, respectively. He is currently a Professor with the College of Computer Science and Technology, Jilin University. His research interests are in the areas of computer vision, deep learning, and data mining
\end{IEEEbiography}


\end{document}